\documentclass[table,xcdraw]{article}
\usepackage{microtype}
\usepackage{graphicx}
\usepackage{subfigure}
\usepackage{booktabs}
\usepackage{hyperref}

\usepackage[accepted]{icml2024}
\usepackage{amsmath}
\usepackage{amssymb}
\usepackage{mathtools}
\usepackage{amsthm}
\usepackage[capitalize,noabbrev]{cleveref}
\usepackage[textsize=tiny]{todonotes}

\usepackage{multirow}
\usepackage{xcolor}
\usepackage{mathrsfs}
\usepackage{units}
\usepackage{pifont}
\usepackage{multicol}
\newtheorem{thm}{Theorem}
\newtheorem{lem}{Lemma}
\newtheorem{rmk}{Remark}
\newtheorem{asmp}{Assumption}

\icmltitlerunning{FedBAT: Communication-Efficient Federated Learning via Learnable Binarization}
\begin{document}
\twocolumn[
\icmltitle{FedBAT: Communication-Efficient \\Federated Learning via Learnable Binarization}

\icmlsetsymbol{intern}{*}
\begin{icmlauthorlist}
\icmlauthor{Shiwei Li}{hust,intern}
\icmlauthor{Wenchao Xu}{polyu}
\icmlauthor{Haozhao Wang}{hust}
\icmlauthor{Xing Tang}{fit}
\icmlauthor{Yining Qi}{hust}
\icmlauthor{Shijie Xu}{fit}
\icmlauthor{Weihong Luo}{fit}
\icmlauthor{Yuhua Li}{hust}
\icmlauthor{Xiuqiang He}{fit}
\icmlauthor{Ruixuan Li}{hust}
\end{icmlauthorlist}

\icmlaffiliation{hust}{Huazhong University of Science and Technology, Wuhan, China}
\icmlaffiliation{polyu}{The Hong Kong Polytechnic University, Hong Kong, China}
\icmlaffiliation{fit}{FiT, Tencent, Shenzhen, China}

\icmlcorrespondingauthor{Haozhao Wang}{hz\_wang@hust.edu.cn}
\icmlcorrespondingauthor{Xing Tang}{xing.tang@hotmail.com}
\icmlcorrespondingauthor{Ruixuan Li}{rxli@hust.edu.cn}

\icmlkeywords{Federated Learning, Communication Compression, Binarization}
\vskip 0.3in
]
\printAffiliationsAndNotice{\textsuperscript{*}This work was done when Shiwei Li worked as an intern at FiT, Tencent.}  

\begin{abstract}
Federated learning is a promising distributed machine learning paradigm that can effectively exploit large-scale data without exposing users' privacy. However, it may incur significant communication overhead, thereby potentially impairing the training efficiency. To address this challenge, numerous studies suggest binarizing the model updates. Nonetheless, traditional methods usually binarize model updates in a post-training manner, resulting in significant approximation errors and consequent degradation in model accuracy. To this end, we propose \textbf{Federated Binarization-Aware Training (FedBAT)}, a novel framework that directly learns binary model updates during the local training process, thus inherently reducing the approximation errors. FedBAT incorporates an innovative binarization operator, along with meticulously designed derivatives to facilitate efficient learning. In addition, we establish theoretical guarantees regarding the convergence of FedBAT. Extensive experiments are conducted on four popular datasets. The results show that FedBAT significantly accelerates the convergence and exceeds the accuracy of baselines by up to 9\%, even surpassing that of FedAvg in some cases. 
\end{abstract}
\section{Introduction}\label{sec:intro}

\begin{figure*}[t]
\centering
\includegraphics[scale=0.66]{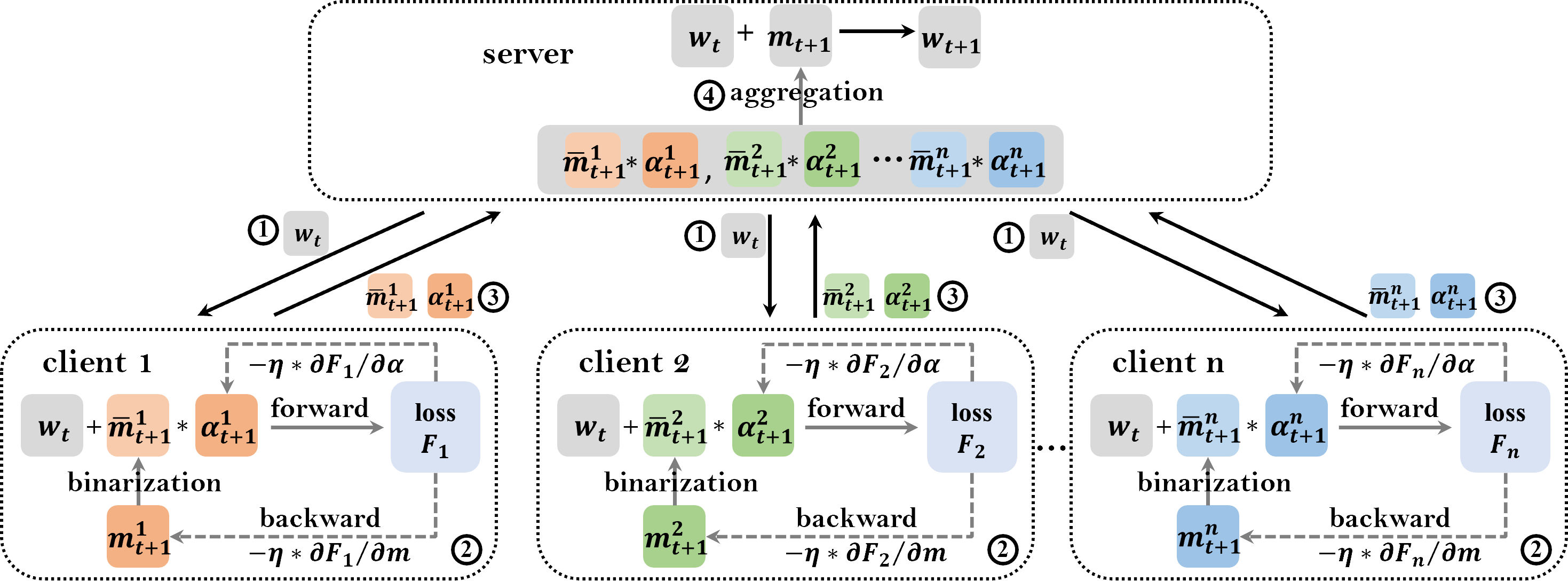}
\caption{An illustration of the $t$-{th} round within the FedBAT framework. \ding{172} downlink: the server sends model parameters $\mathbf{w}_{t}$ to clients; \ding{173} local training: clients train the model updates ($\boldsymbol{m}_{t+1}$ and $\mathbf{\alpha}_{t+1}$) via learnable binarization; \ding{174} uplink: clients upload their binary model updates ($\bar{\boldsymbol{m}}_{t+1}$ and $\mathbf{\alpha}_{t+1}$) to the server. \ding{175} model aggregation: the server aggregates binary model updates to generate $\mathbf{w}_{t+1}$.} 
\label{fig:fedbat}
\end{figure*}

Federated learning (FL)~\cite{fedavg} stands out as a promising distributed machine learning paradigm designed to safeguard data privacy. It enables distributed clients to collaboratively train a global model without sharing their local data, thus can protect user data privacy. In the classic FL system, a central server initially broadcasts the global model to several clients, who then train it using their respective datasets. After local training, the clients send their model parameters or model updates back to the server, who aggregates them to create a new global model for the next round of training. This process is repeated for several rounds until the global model converges. 

Despite FL's success in preserving local data privacy, the iterative transmission of model parameters introduces considerable communication overhead, adversely affecting training efficiency. Specifically, the communication occurs in two stages: the uplink, where clients send model updates to the server, and the downlink, where the server broadcasts the global model to clients. As suggested by \citet{dadaquant}, the uplink typically imposes a tighter bottleneck than the downlink, particularly due to the global mobile upload bandwidth being less than one fourth of the download bandwidth. Therefore, in this paper, we seek to compress the uplink communication, as much research~\cite{fedpaq,fedpm,zsign} has done.

SignSGD~\cite{signsgd} is an effective binarization technique for reducing communication volume. It was originally proposed to communicate only the signs of gradients in distributed training systems. Recently, it has also been naturally applied to binarize model updates in FL~\cite{bayesian-signsgd}. Specifically, it binarizes model updates $\boldsymbol{m}\in\mathbb{R}^d$ as $\boldsymbol{\hat{m}}=\alpha\cdot \boldsymbol{\bar{m}} = \alpha\cdot\text{sign}(\boldsymbol{m})$, where $\alpha$ denotes the magnitude of binary values, termed the step size and typically tuned as a hyperparameter. SignSGD can efficiently compress the uplink communication by a factor of 32. However, the binarized model updates inevitably contains approximation errors in comparison to the original updates, which affects both the convergence speed and the model accuracy.

To improve the performance of SignSGD, subsequent research has mainly explored from two aspects, including error feedback methods~\cite{ef-signsgd} and stochastic sign methods~\cite{noisy-signsgd,stoc-signsgd}. EF-SignSGD~\cite{ef-signsgd} aims to compensate for the errors introduced by binarization. Notably, EF-SignSGD adds the binarized errors from the previous round onto the current model updates before conducting another binarization. 
On the other hand, stochastic sign methods aim to alleviate the bias estimation of SignSGD, namely $\mathbb{E}[\alpha\cdot\text{sign}(\boldsymbol{m})]\neq\boldsymbol{m}$. The key idea here is to perturb local model updates with random noise. Noisy-SignSGD~\cite{noisy-signsgd} adds Gaussian noise $\zeta\sim N(0,\sigma^2)$ to the model update and then performs binarization, where $\sigma$ is an adjustable standard deviation. 
Similarly, for a model update $\boldsymbol{m}$, Stoc-SignSGD~\cite{stoc-signsgd} adds uniform noise $\zeta\sim U(-\Vert\boldsymbol{m}\Vert, \Vert\boldsymbol{m}\Vert)$ to $\boldsymbol{m}$ before binarization. In this way, each element $\boldsymbol{m}_i$ will be binarized into +1 with a probability of ($\nicefrac{1}{2}+\nicefrac{\boldsymbol{m}_i}{2 \Vert\boldsymbol{m}\Vert}$) and into -1 with a probability of ($\nicefrac{1}{2}-\nicefrac{\boldsymbol{m}_i}{2 \Vert\boldsymbol{m}\Vert}$). 

The above research may differ in their motivations and specific solutions, nevertheless, they share a fundamental similarity. Before binarization, a compensation term or disturbance term will be superimposed on the model updates to be binarized, which is the binarized errors from the previous round in EF-SignSGD, Gaussian noise in Noisy-SignSGD, and uniform noise in Stoc-SignSGD. Although applying these methods in FL can indeed enhance the performance of SignSGD (see Section~\ref{sec:overall}), it is crucial to acknowledge that they still suffer from two main drawbacks. 
On one hand, the above methods still binarize model updates in a post-training manner as SignSGD does. Binarized errors are introduced solely after the training phase, depriving them of the opportunity to be optimized during local training. 
On the other hand, the step size is usually kept as a hyperparameter, necessitating significant human effort to tune it for various applications. However, even with careful tuning, the resulting performance may still fall short of the optimum. 

In this paper, we aim to solve these two issues by exploiting the local training process of FL. Specifically, our primary goal is to equip clients with the capability of directly learning binary model updates and the corresponding step size. To this end, we propose a novel training paradigm, termed \textbf{Federated Binarization-Aware Training (FedBAT)}. The illustration of each round within FedBAT is presented in Figure~\ref{fig:fedbat}. 
The key idea of FedBAT is to binarize the model update with the step size during the forward propagation. This entails calculating the output loss based on the binarized model updates. It provides an opportunity to compute gradients for both the binarized model updates and the step size, facilitating their subsequent optimization. However, the derivative of the vanilla binarization operator is always zero, making the gradient descent algorithm infeasible. To solve this issue, we further introduce a learnable binarization operator with well-designed derivatives. 

The main contributions can be summarized as follows: 
\begin{itemize}
\item We analyze that the drawbacks of existing federated binarization methods lie in their post-training manner. This opens new avenues for binarization in FL.
\item Based on our analysis, we propose FedBAT to learn binary model updates during the local training process of FL. For this, a novel binarization operator is employed with well-designed derivatives. 
\item Theoretically, we establish convergence guarantees for FedBAT, demonstrating a comparable convergence rate to its uncompressed counterpart, the FedAvg. 
\item Experimentally, we validate FedBAT on four popular datasets. The experimental results show that FedBAT can significantly improve the convergence speed and test accuracy of existing binarization methods, even surpassing the accuracy of FedAvg in some cases. The code is available at \url{https://github.com/Leopold1423/fedbat-icml24}.
\end{itemize}

\section{Preliminaries}
FL involves $N$ clients connecting to a server. The general goal of FL is to train a global model by multiple rounds of local training on each client's local dataset. 
Denoting the objective function of the $k$-th client as $F_k$, FL can be formulated as 
\begin{equation}
    \min_{\mathbf{w}}F(\mathbf{w}) = \sum _{k=1}^{N} p_k F_k(\mathbf{w}), 
\end{equation}
where $p_k$ is the proportion of the $k$-{th} client's data to all the data of the $N$ clients. 
FedAvg~\cite{fedavg} is a widely used FL algorithm. In the $t$-{th} round, the server sends the model parameters $\mathbf{w}_{t}$ to several randomly selected $K$ clients, denoted as $\mathbb{S}_{t}$. 
Each selected client performs certain steps of local training and sends the model update $\mathbf{w}_{t+1}^k-\mathbf{w}_{t}$ back to the server. The server aggregates these updates to generate a new global model as follows: 
\begin{equation}\label{eq:fedavg}
    \mathbf{w}_{t+1} = \mathbf{w}_{t} + \sum_{k \in \mathbb{S}_{t}} p'_k(\mathbf{w}_{t+1}^k-\mathbf{w}_{t}), 
\end{equation}
where $p'_k=\nicefrac{p_k}{\sum _{\mathbb{S}_{t}}p_k}$ denotes the proportion of the $k$-th client's data to all the data used in the $t$-th round.

In this paper, we employ SignSGD and its aforementioned variants to compress the uplink communication in FL. Each client only needs to send the signs of its model updates to the server. Note that the signs undergo a encoding $\{-1, 1\}\rightarrow\{0, 1\}$ before transmission, and are decoded back after transmission. For simplicity, we omit this process in the following pages. Therefore, Eq.\ref{eq:fedavg} can be rewritten as
\begin{equation}\label{eq:signsgd}
    \mathbf{w}_{t+1} = \mathbf{w}_{t} + \sum _{k\in\mathbb{S}_{t}} p'_k \alpha \cdot \text{sign}(\mathbf{w}_{t+1}^k-\mathbf{w}_{t}). 
\end{equation}

\section{Methodology}
\subsection{Motivations and Objectives}
In the $t$-th round, the objective of local training for the $k$-th client can be formulated as
\begin{equation}
    \min _{\boldsymbol{m}_{t+1}^k} F_k(\mathbf{w}_{t}+\boldsymbol{m}_{t+1}^k),
\end{equation}
where $F_k$ is the loss function of the $k$-th client. $\mathbf{w}_{t}$ is the global model parameter in the $t$-{th} round and $\boldsymbol{m}_{t+1}^k$ represents the model updates to be learned. Binarization is not considered as a constraint during local training but is only performed after obtaining the final updates $\boldsymbol{m}_{t+1}^k$. 

The binarized updates ${\hat{\boldsymbol{m}}_{t+1}^k}$ inevitably contain errors compared to the original updates $\boldsymbol{m}_{t+1}^k$, thus reducing the model accuracy and slowing down the convergence speed. A natural motivation is to correct or compensate for the binarized errors during the local training process. 
In light of this, we try to introduce binarization of model updates into local training, thereby directly learning binarized model updates and their corresponding step size. Specifically, the objective of local training can be formulated as
\begin{equation}
    \min _{\boldsymbol{m}_{t+1}^k, \alpha_{t+1}^k} F_k(\mathbf{w}_{t}+\mathcal{S}(\boldsymbol{m}_{t+1}^k, \alpha_{t+1}^k)),
\end{equation}
where $\mathcal{S}(\boldsymbol{m}_{t+1}^k, \alpha_{t+1}^k)$ are the binary model updates to be learned. $\mathcal{S}$ is a binarization operator. The model updates $\boldsymbol{m}_{t+1}^k$ and the step size $\alpha_{t+1}^k$ are learnable parameters. 

Next, we will introduce the learnable binarization operator $\mathcal{S}$ in Section~\ref{sec:learnable}, and then present the detailed pipeline of FedBAT in Section~\ref{sec:fedbat}. Theoretical guarantees on the convergence of FedBAT will be provided in the next section.

\subsection{Learnable Binarization}\label{sec:learnable}
We first define a binarization operator as follows:
\begin{equation}
\mathcal{S}(x, \alpha) = \left\{
\begin{array}{ll}
    \alpha    & x >\alpha,  \\ 
    \mathcal{S}'(x,\alpha)    & -\alpha \leq x \leq \alpha,\\ 
    -\alpha  & x < -\alpha,
\end{array}
\right. \label{eq:binary_operator}
\end{equation}
where $\mathcal{S}'(x, \alpha)$ is a stochastic binarization with  uniform noise $\zeta\sim U(0,1)$ added as follows:
\begin{equation}
\begin{aligned}
\mathcal{S}'(x, \alpha) &= \alpha (2\lfloor \nicefrac{\alpha+x}{2\alpha} + \zeta\rfloor-1) \\
&= \left\{
\begin{array}{ll}
    \alpha    & \text{w.p.} \quad \nicefrac{\alpha+x}{2\alpha},\\ 
    -\alpha   & \text{w.p.} \quad \nicefrac{\alpha-x}{2\alpha}. \\
\end{array}
\right. \\
\end{aligned}
\label{eq:s}
\end{equation}
However, the floor function $\lfloor x\rfloor$ returns the largest integer not exceeding $x$, and its derivative is always zero, which makes gradient descent infeasible. To solve this issue, we adopt Straight-through Estimator (STE)~\cite{ste} to treat the floor function as an identity map during backpropagation, that is to say its derivative is equal to 1. 
Recent work has demonstrated that STE works as a first-order approximation of the gradient and affirmed its efficacy~\cite{reinmax}. Therefore, the gradient of $x$ can be estimated by: 
\begin{equation}
\nicefrac{\partial\mathcal{S}}{\partial x} = \left\{
\begin{array}{ll}
    0    & x >\alpha,\\ 
    1    & -\alpha \leq x \leq \alpha,\\ 
    0    & x < -\alpha,
\end{array}
\right. \label{eq:db}
\end{equation}
and the gradient of $\alpha$ can be estimated by: 
\begin{equation}
\nicefrac{\partial\mathcal{S}}{\partial \alpha} = \left\{
\begin{array}{ll}
    1    & x >\alpha,\\ 
    2\lfloor \nicefrac{\alpha+x}{2\alpha} + \zeta\rfloor-\nicefrac{x+\alpha}{\alpha}    & -\alpha \leq x \leq \alpha,\\ 
    -1    & x < -\alpha,
\end{array}
\right. \label{eq:dalpha}
\end{equation}

It is worth noting that $\alpha$ represents the step size and is supposed to be a positive number, otherwise the meaning of Eq.\ref{eq:binary_operator} will be wrong. To restrict the value of $\alpha$ to be positive, we calculate $\alpha$ as follows:
\begin{equation}
\alpha = \alpha' e^{\rho\alpha_e}, \label{eq:alpha}
\end{equation}
where $\alpha_e$ becomes the learnable step size, initialized to zero. $\alpha'$ is the initial value of $\alpha$ and $\rho$ is a hyperparameter that
regulates the pace of the optimization process for $\alpha_e$. 

So far, we have developed a derivable binarization operator, which can be seamlessly integrated into a neural network. Let $\mathbf{x}$ denotes a layer of parameters within the network, and $\alpha$ denotes the corresponding step size calculated by Eq.\ref{eq:alpha}. 
During forward propagation, $\mathbf{x}$ will undergo element-wise binarization using Eq.\ref{eq:binary_operator}, where $\alpha$ is shared by all elements in $\mathbf{x}$. 
During backpropagation, $\mathbf{x}$ and $\alpha$ will be optimized by the gradients calculated from Eq.\ref{eq:db} and Eq.\ref{eq:dalpha}. 

\subsection{Federated Binarization-Aware Training}\label{sec:fedbat}
Here, we integrate the learnable binarization operator into local training of FL. In contrast to SignSGD, this allows clients to consider binarized errors and make corrections when optimizing locally. We refer to the proposed training framework as Federated Binarization-Aware Training (FedBAT), and the pipeline is shown in Algorithm~\ref{alg:fedbat}.

In FedBAT, the operating procedures of the server is the same as that in FedAvg. As described in lines 4-8, at the beginning of each round, the server sends global model parameters to several randomly selected clients. It subsequently collects the binarized model updates returned by the clients to calculate new global model parameters as follows:
\begin{equation}\label{eq:bat_aggregate}
    \mathbf{w}_{t+1} = \mathbf{w}_{t} + \sum_{k\in\mathbb{S}_{t}} p'_k \hat{\boldsymbol{m}}_{t+1}^k = \mathbf{w}_{t} + \sum_{k\in\mathbb{S}_{t}} p'_k\alpha_{t+1}^k \bar{\boldsymbol{m}}_{t+1}^k,
\end{equation}
where $\hat{\boldsymbol{m}}_{t+1}^k$ and $\alpha_{t+1}^k$ are the binarized model updates and the step size of the $k$-{th} client. Note that the model update of each layer within the model has a unique step size, thus, Eq.\ref{eq:bat_aggregate} is actually performed layer-wise. For simplicity, we omit the representations of the different layers.

\begin{algorithm}[tb]
\caption{\textbf{Federated Binarization-Aware Training}}\label{alg:fedbat}
\begin{algorithmic}[1]
\STATE {\bfseries Input:} the iteration rounds $R$; the local steps $\tau$; the local warm-up ratio $\phi$, the coefficient of the step size $\rho$.
\FUNCTION{\textbf{server}($R$)}
\STATE Initialize global model parameters with $\mathbf{w}_0$.
\FOR{$t=0$ {\bfseries to} $R-1$}
\STATE Send global model parameters $\mathbf{w}_{t}$ to clients.
\STATE Get model updates ($\bar{\boldsymbol{m}}_{t+1}$, $\alpha_{t+1}$) from clients.
\STATE Aggregate binary model updates by Eq.\ref{eq:bat_aggregate}.
\ENDFOR
\ENDFUNCTION

\FUNCTION {\textbf{client}($\mathbf{w}_{t}$, $\tau$, $\phi$, $\rho$)}
\STATE Initialize local model parameters with $\mathbf{w}_{t}$.
\STATE Initialize local model updates $\boldsymbol{m}_{t}$ with zeros. 
\FOR{$s=0$ {\bfseries to} $\tau-1$}
\IF {$s < \lfloor \phi\tau \rfloor$}
\STATE Run full-precision training by Eq.\ref{eq:fpt}.
\ELSIF {$s = \lfloor \phi\tau \rfloor$}
\STATE Initialize the step size layer-wise by Eq.\ref{eq:alpha_init}.
\STATE Run binarization-aware training by Eq.\ref{eq:bat}.
\ELSE
\STATE Run binarization-aware training by Eq.\ref{eq:bat}.
\ENDIF
\ENDFOR
\STATE $\bar{\boldsymbol{m}}_{t+1}^k$, $\alpha_{t+1}^k = \bar{\boldsymbol{m}}_{t,\tau-1}^k$, $\alpha_{t,\tau-1}^k$.
\STATE Send binary model updates ($\bar{\boldsymbol{m}}_{t+1}^k$, $\alpha_{t+1}^k$) to server.
\ENDFUNCTION
\end{algorithmic}
\end{algorithm}

Before introducing the local training procedures in FedBAT, we highlight the variances in the local model architectures. We reiterate that FedBAT is achieved by performing the binarization operator defined in Eq.\ref{eq:binary_operator} on the model updates during local training. However, there is no explicit representation of model updates within the vanilla model. Therefore, we allow local models to maintain an extra copy of model parameters to represent model updates, denoted as $\boldsymbol{m}$. In addition, the step size $\alpha$ of the update within each layer is also kept as a learnable parameter.

The local training process comprises two distinct stages: full-precision training and binarization-aware training. In the $t$-{th} round, as depicted in lines 11-12, each client loads the global model parameters $\mathbf{w}_{t}$ and initialize its model updates $\boldsymbol{m}_{t}$ with zeros. Given that the model updates commence as zeros, learning to binarize them becomes more challenging. Therefore, to secure more precise initialization values for the model updates, we initially conduct training without binarization, that is the full-precision training where model updates will be optimized as follows:
\begin{equation} ~\label{eq:fpt}
\begin{aligned}
    \boldsymbol{m}_{t,s+1}^k &= \boldsymbol{m}_{t,s}^k -\eta_t \nicefrac{\partial F_k(\mathbf{w}_t^k + \boldsymbol{m}_{t,s}^k)}{\partial \boldsymbol{m}_{t,s}^k}.
\end{aligned}
\end{equation}
A warm-up ratio $\phi$ is defined to denote the proportion of full-precision training to the entire local training. 
After the full-precision training, $\alpha'$ and $\alpha_e$ defined in Eq.\ref{eq:alpha} will be initialized for each layer of the $k$-th client as follows: 
\begin{equation}\label{eq:alpha_init}
    (\alpha')^k_l =  \nicefrac{\Vert(\boldsymbol{m})^k_l\Vert_1}{d^k_l}, \quad (\alpha_e)^k_l = 0,
\end{equation}
where $(\boldsymbol{m})^k_l\in \mathbb{R}^{d^k_l}$ represents the model update of the $l$-th layer in the $k$-th client. Next, FedBAT performs binarization-aware training, optimizing the step size and the binarized model update using Eq.\ref{eq:db} and Eq.\ref{eq:dalpha} as follows:
\begin{equation}\label{eq:bat}
\begin{aligned}
    \boldsymbol{m}_{t,s+1}^k &= \boldsymbol{m}_{t,s}^k -\eta_t \nicefrac{\partial F_k(\mathbf{w}_t^k + \hat{\boldsymbol{m}}_{t,s}^k)}{\partial \boldsymbol{m}_{t,s}^k},\\
    \alpha_{t,s+1}^k &= \alpha_{t,s}^k -\eta_t \nicefrac{\partial F_k(\mathbf{w}_t^k + \hat{\boldsymbol{m}}_{t,s}^k)}{\partial \alpha_{t,s}^k}, \\
    \hat{\boldsymbol{m}}_{t,s+1}^k &= \mathcal{S}(\boldsymbol{m}_{t,s+1}^k, \alpha_{t,s+1}^k).
\end{aligned}
\end{equation}
After local training, the $k$-{th} client will send its binarized model update $\bar{\boldsymbol{m}}_{t+1}^k$ and the step size $\alpha_{t+1}^k$ to the server.

\section{Convergence Analysis}\label{sec:convergence}
In this section, we provide theoretical guarantees for the convergence of FedBAT, while considering the data heterogeneity within FL. For simplicity, we focus on the case where the warm-up ratio $\phi$ is zero, meaning that the warm-up training, as defined by Eq.\ref{eq:fpt}, is not performed. Furthermore, to ensure the unbiased property of the binarization operator $\mathcal{S}$ in Eq.\ref{eq:binary_operator}, we set the step size as $\alpha_{t,s}^k=\Vert \boldsymbol{m}_{t,s}^k\Vert_\infty$ in each step instead of optimizing it during local training. Then we give the following notations and assumptions. 

\textbf{Notations.} Let $F^*$ and $F_k^*$ be the minimum values of $F$ and $F_k$, respectively, then $\Gamma = F^* - \sum_{k=1}^N p_k F_k^* $ can be used to quantify the degree of data heterogeneity. $\tau$ is the number of local steps. In Section~\ref{sec:fedbat}, the subscripts $t\in[R]$ and $s\in[\tau]$ are used to represent the serial number of global rounds and local iterations, respectively. In the following analysis, we will only use the subscript $t$ to represent the cumulative number of iteration steps in the sense that $t\in[T] ,T =R\tau$.

\begin{asmp}\label{asmp:lsmooth}
$F_1,...,F_N$ are all $L$-smooth: for $\mathbf{w}$ and $\mathbf{v}$, $F_k(\mathbf{v})\leq F_k(\mathbf{w}) + (\mathbf{v}-\mathbf{w})^T\nabla F_k(\mathbf{w}) + \frac{L}{2}\Vert \mathbf{v} -\mathbf{w}\Vert^2$.
\end{asmp}
\begin{asmp}\label{asmp:ustrong}
$F_1,...,F_N$ are $u$-strongly convex: for all $\mathbf{w}$ and $\mathbf{v}$, $F_k(\mathbf{v})\geq F_k(\mathbf{w}) + (\mathbf{v}-\mathbf{w})^T\nabla F_k(\mathbf{w}) + \frac{\mu}{2}\Vert \mathbf{v} -\mathbf{w}\Vert^2$.
\end{asmp}
\begin{asmp}\label{asmp:gvaricane}
Let $\xi_t^k$ be sampled from the $k$-th client’s local data uniformly at random. The variance of stochastic gradients in each device is bounded: $\mathbb{E} \Vert \nabla F_k(\mathbf{w}_t^k, \xi_t^k) - \nabla F_k(\mathbf{w}_t^k)\Vert^2\leq \sigma^2$ for all $k = 1,...,N$.
\end{asmp}
\begin{asmp}\label{asmp:gbound}
The expected squared norm of stochastic gradients is uniformly bounded, i.e., $\mathbb{E} \Vert \nabla F_k(\mathbf{w}_t^k, \xi_t^k) \Vert^2\leq G^2$ for all $k = 1,...,N$ and $t = 1,...,T$.
\end{asmp}
\begin{asmp}\label{asmp:svariance}
The variance of binarization $\mathcal{S}$ grows with the $l_2$-norm of its argument, i.e., $\mathbb{E} \Vert \mathcal{S}(\mathbf{x}) - \mathbf{x} \Vert \leq q \Vert \mathbf{x} \Vert$.
\end{asmp}

Assumptions \ref{asmp:lsmooth}-\ref{asmp:gbound} are commonplace in standard optimization analyses~\cite{sparse_sgd,model_avg,fedavg_convergence}.
The condition in Assumption~\ref{asmp:svariance} is satisfied with many compression schemes including the binarization operator $\mathcal{S}$ as defined in Eq.\ref{eq:binary_operator}. Assumption~\ref{asmp:svariance} is also used in \cite{ef-signsgd, fedpaq} to analyze the convergence of federated algorithms.
Theorems \ref{thm:full} and \ref{thm:part} show the convergence of FedBAT under the strongly convex assumption with full and partial device participation, respectively. The convergence of FedBAT under the non-convex assumption with partial device participation is shown in Theorems \ref{thm:non-convex}. All proofs are provided in Appendix.

\begin{thm}\label{thm:full}
Let Assumptions \ref{asmp:lsmooth}-\ref{asmp:svariance} hold and $L, \mu, \sigma, G, q$ be defined therein. Choose $\kappa = \frac{L}{\mu}, \gamma = \max\{8\kappa, \tau\}-1$ and the learning rate $\eta_t = \frac{2}{\mu(\gamma+t)}$. Then FedBAT with full device participation satisfies
\begin{equation}
\begin{aligned}
    \mathbb{E}[F(\mathbf{w}_T)] - F^* \leq \frac{\kappa}{\gamma+T}(\frac{2B}{\mu}+\frac{\mu(\gamma+1)}{2}\mathbb{E}\Vert\mathbf{w}_1 - \mathbf{w}^*\Vert^2),
\end{aligned}
\end{equation}
where $B=\sum_{k=1}^N p_k^2\sigma^2 + 6L\Gamma + 8(1+q^2)(\tau-1)^2G^2 + 4\sum_{k=1}^N p_k^2q^2\tau^2G^2$.
\end{thm}
\begin{thm}\label{thm:part}
Let Assumptions \ref{asmp:lsmooth}-\ref{asmp:svariance} hold and $L, \mu, \sigma, G, q$ be defined therein. Let $\kappa, \gamma, \eta_t$ be defined in Theorem \ref{thm:full}. Assuming that $K$ devices are randomly selected to participate in each round of training and their data is balanced in the sense that $p_1=...=p_N=\frac{1}{N}$. Then the same bound in Theorem \ref{thm:full} holds if we redefine the value of $B$ to 
$B=\frac{\sigma^2}{N} + 6L\Gamma + 8(1+q^2)(\tau-1)^2G^2 + 4\frac{q^2(N-1)+N-K}{K(N-1)}\tau^2G^2$.
\end{thm}

\begin{rmk}
By setting $K=N$, Theorem \ref{thm:part} transforms into Theorem \ref{thm:full}. By setting $q = 0$, Theorem \ref{thm:full} and \ref{thm:part} are equivalent to the analysis of FedAvg in \cite{fedavg_convergence}. 
By setting $K=N$ and $\tau=1$, Theorem \ref{thm:full} and \ref{thm:part} recovers the convergence rate of Stoc-SignSGD~\cite{stoc-signsgd} when used in distributed training. By setting $K=N, \tau=1$ and $q = 0$, Theorem \ref{thm:full} and \ref{thm:part} recovers the convergence rate of vanilla SGD, i.e., $\mathcal{O}(\frac{1}{T})$ for strongly-convex losses. 
\end{rmk}

\begin{thm}\label{thm:non-convex}
Let Assumptions 1 and 3-5 hold, i.e., without the convex assumption, and $L$, $\sigma$, $G$, $q$ be defined therein. Assume the learning rate is set to $\eta = \frac{1}{L\sqrt{T}}$ and the local dataset is balanced, then the following first-order stationary condition holds for FedBAT with partial device participation
\begin{equation}
    \frac{1}{T}\sum _{t=0}^{T-1} \mathbb{E}\Vert \nabla F(\mathbf{w}_t)\Vert^2 \leq \frac{2L(F(\mathbf{w}_0) - F^* + \Gamma)}{\sqrt{T}} + \frac{P}{\sqrt{T}} + \frac{Q}{T},
\end{equation}
where $P = \frac{\sigma^2}{N} + 4\frac{q^2(N-1)+N-K}{K(N-1)}\tau^2G^2$ and $Q=4(1+q^2)(\tau-1)^2G^2$. 
\end{thm}

\begin{rmk}
Under the conditions of Theorems \ref{thm:full}-\ref{thm:non-convex}, the convergence rate of both FedBAT and FedAvg ($q=0$) is $\mathcal{O}(\frac{1}{T})$ in the strongly convex setting, and $\mathcal{O}(\frac{1}{T}) + \mathcal{O}(\frac{1}{\sqrt{T}})$ in the non-convex setting.
\end{rmk}
\begin{rmk}
For ease of analysis, FedBAT is discussed in the case where the step size is set as $\alpha^{t,s}_i=\Vert \boldsymbol{m}^{t,s}_i\Vert_\infty$ without optimization. However, learning the step size shall be able to achieve smaller value of $q$ and enhance the performance of FedBAT. Empirically, we show in Section~\ref{sec:exp} that optimizing the step size as defined in Eq.\ref{eq:alpha_init}-\ref{eq:bat} achieves better accuracy. 
\end{rmk}

\section{Experiments}\label{sec:exp}
\begin{figure*}[!ht]
\centering
\subfigure[CIFAR-100, IID]{
\begin{minipage}[h]{0.31\textwidth}
\centering
\includegraphics[scale=0.36]{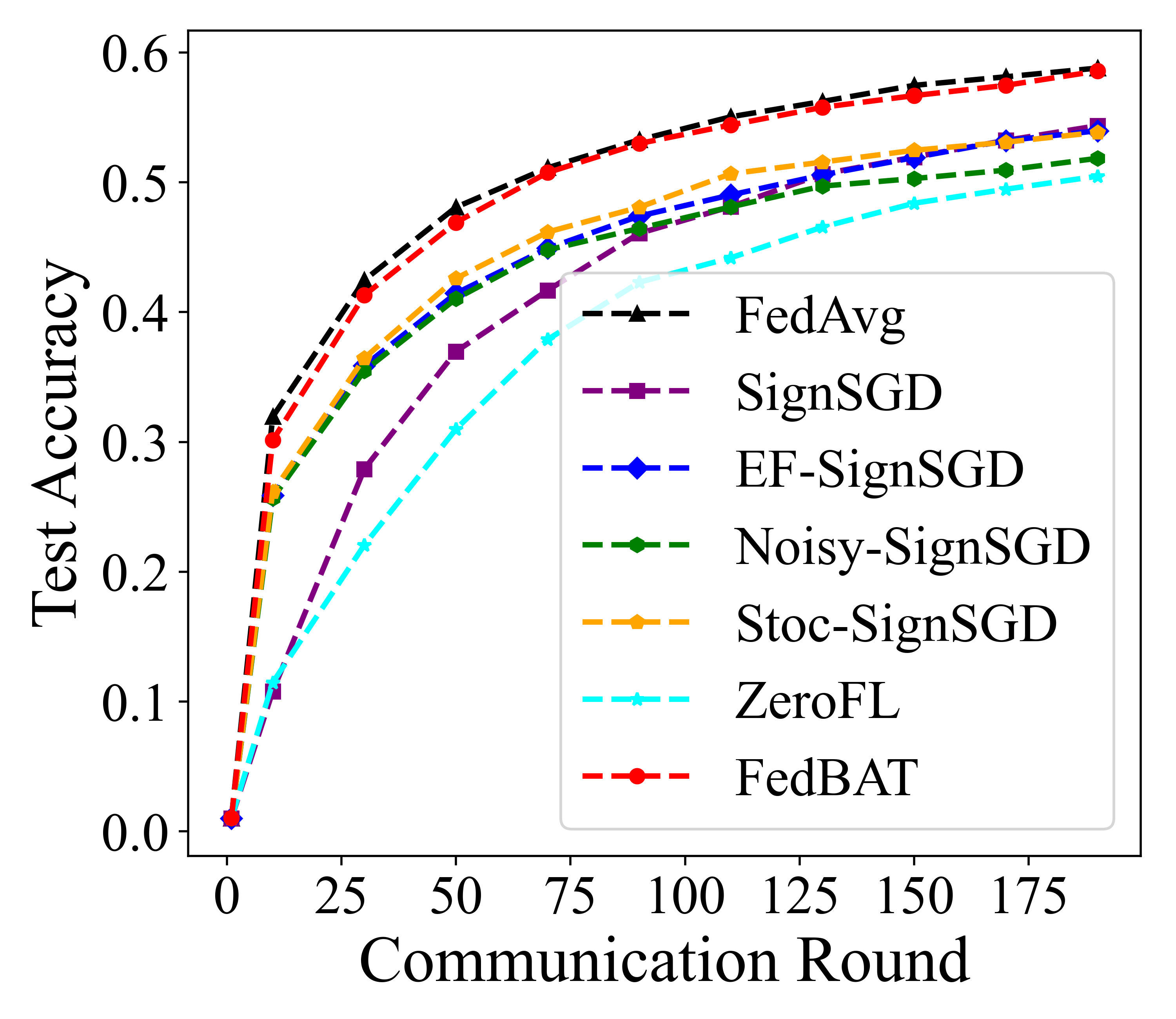}
\end{minipage}
}
\subfigure[CIFAR-100, Non-IID-1]{
\begin{minipage}[h]{0.31\textwidth}
\centering
\includegraphics[scale=0.36]{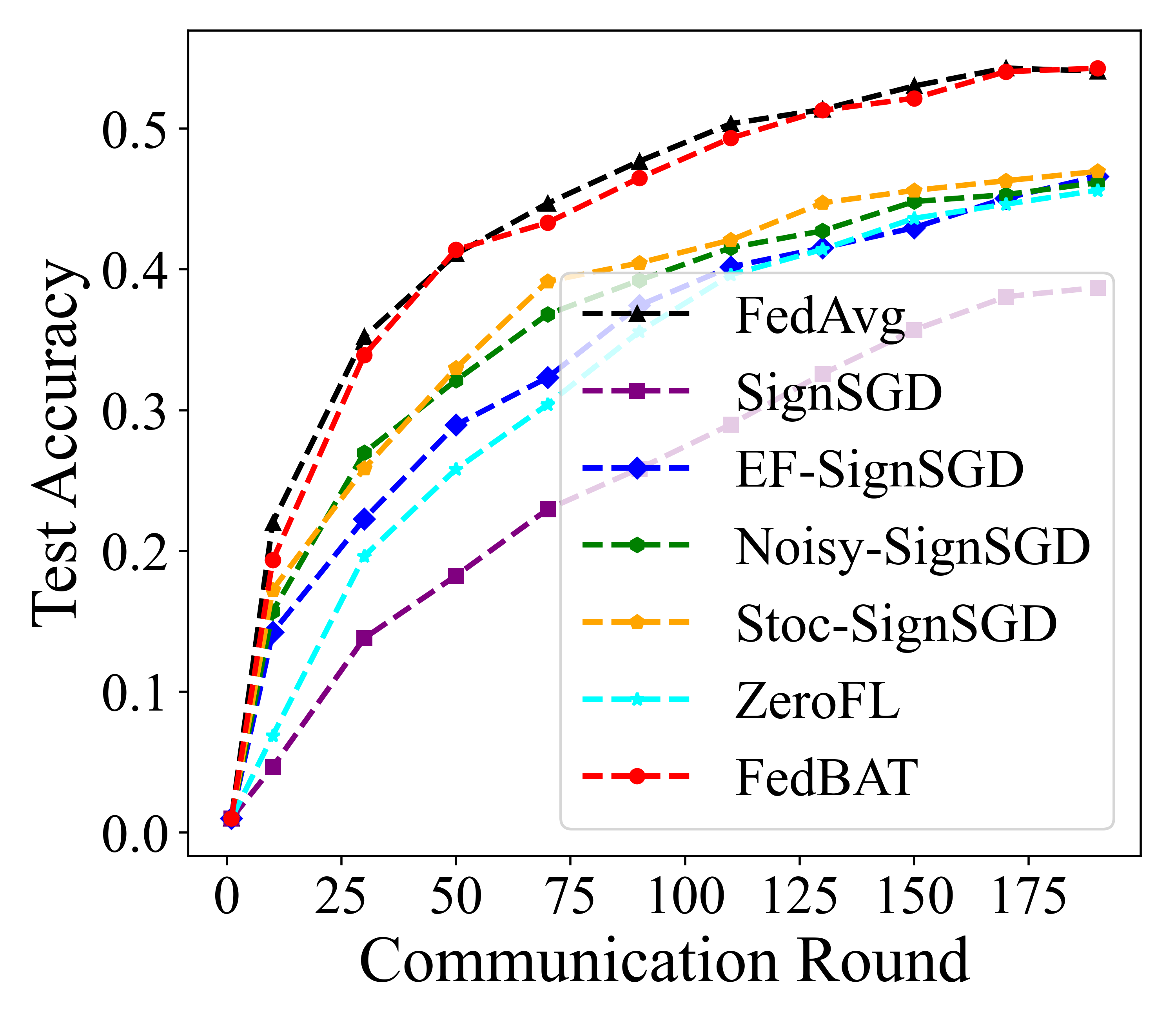}
\end{minipage}
}
\subfigure[CIFAR-100, Non-IID-2]{
\begin{minipage}[h]{0.31\textwidth}
\centering
\includegraphics[scale=0.36]{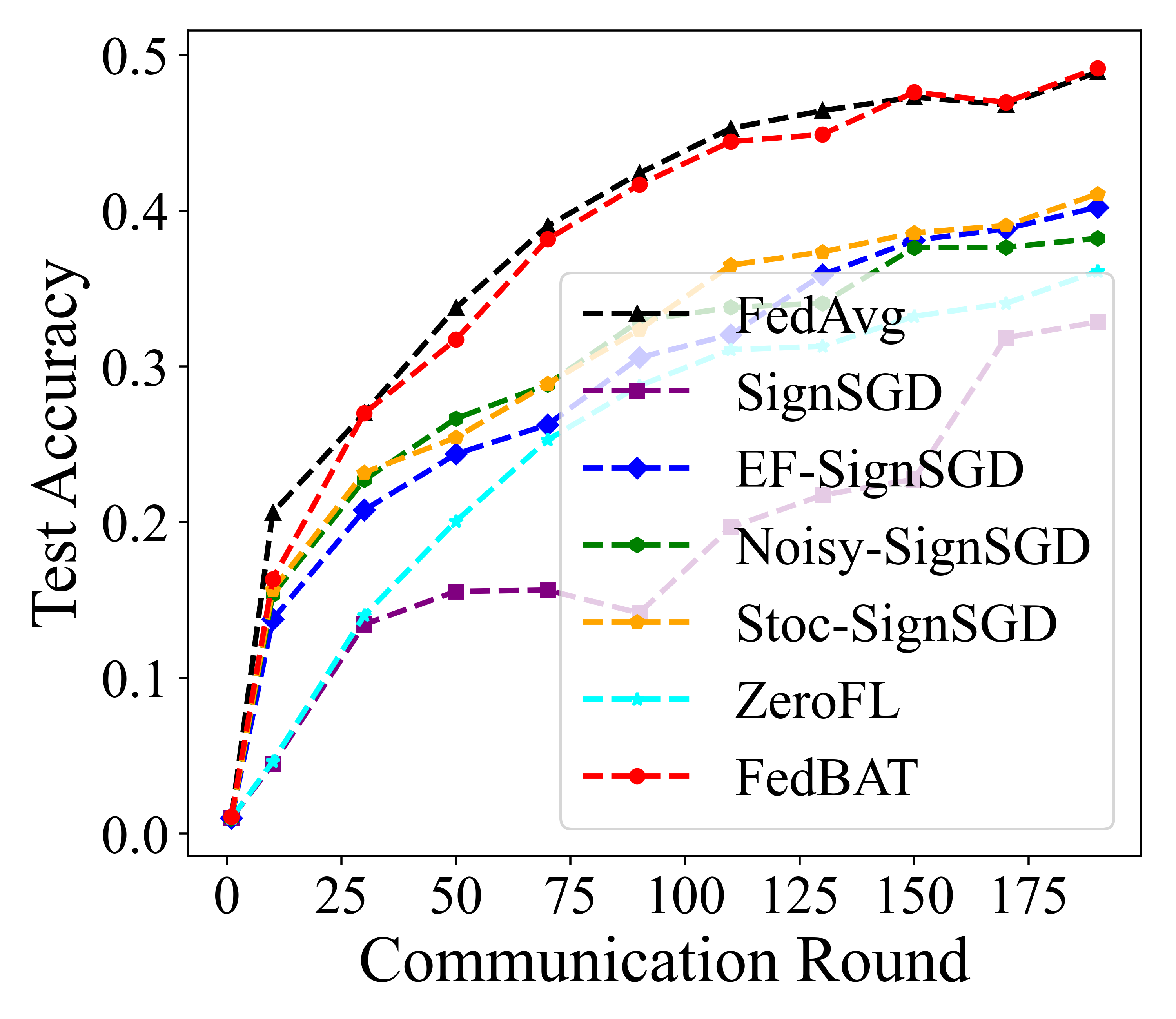}
\end{minipage}
}
\caption{Convergence curves of FedBAT and baselines on CIFAR-100 with 100 clients.}\label{fig:converge}
\end{figure*}
\begin{table*}[ht]
\renewcommand{\arraystretch}{1.0}
\centering
\caption{The test accuracy of all methods on four datasets. The best accuracy is bolded and the next best accuracy is underlined.}\label{tb:performance}
\scalebox{0.75}{
\begin{tabular}{lcccccc}
\toprule[1pt]

& \multicolumn{3}{c}{$N=30$} & \multicolumn{3}{c}{$N=100$} \\ \cmidrule(r){2-4} \cmidrule(r){5-7}

& \makebox[0.13\textwidth][c]{IID}  & \makebox[0.13\textwidth][c]{Non-IID-1} & \makebox[0.13\textwidth][c]{Non-IID-2} & \makebox[0.13\textwidth][c]{IID}  & \makebox[0.13\textwidth][c]{Non-IID-1} & \makebox[0.13\textwidth][c]{Non-IID-2}  \\ 

\multicolumn{7}{c}{\cellcolor[HTML]{C0C0C0}FMNIST with CNN} \\  
FedAvg~\cite{fedavg}                & \textbf{92.5} ($\pm$ 0.1) & \underline{90.7} ($\pm$ 0.1) & \underline{88.9} ($\pm$ 0.2) & \textbf{92.0} ($\pm$ 0.1) & \underline{90.5} ($\pm$ 0.2) & \underline{88.7} ($\pm$ 0.2)  \\
SignSGD~\cite{signsgd}              & 91.3 ($\pm$ 0.1) & 86.5 ($\pm$ 1.0) & 78.9 ($\pm$ 1.3) & 90.3 ($\pm$ 0.1) & 88.2 ($\pm$ 0.2) & 80.5 ($\pm$ 1.0)  \\
EF-SignSGD~\cite{ef-signsgd}        & \underline{92.3} ($\pm$ 0.1) & 90.5 ($\pm$ 0.1) & 88.6 ($\pm$ 0.2) & 91.3 ($\pm$ 0.1) & 89.7 ($\pm$ 0.1) & 87.4 ($\pm$ 0.1)  \\
Noisy-SignSGD~\cite{noisy-signsgd}  & 92.1 ($\pm$ 0.1) & 90.3 ($\pm$ 0.2) & 87.6 ($\pm$ 0.3) & 91.0 ($\pm$ 0.1) & 89.4 ($\pm$ 0.1) & 86.9 ($\pm$ 0.1)  \\
Stoc-SignSGD~\cite{stoc-signsgd}    & 91.7 ($\pm$ 0.1) & 89.5 ($\pm$ 0.2) & 82.6 ($\pm$ 0.8) & 90.6 ($\pm$ 0.1) & 88.5 ($\pm$ 0.2) & 84.8 ($\pm$ 0.8)  \\
ZeroFL~\cite{zerofl}                & 91.0 ($\pm$ 0.1) & 89.3 ($\pm$ 0.3) & 87.4 ($\pm$ 0.2) & 90.2 ($\pm$ 0.1) & 88.8 ($\pm$ 0.2) & 86.6 ($\pm$ 0.1)  \\
FedBAT                               & \textbf{92.5} ($\pm$ 0.1) & \textbf{90.8} ($\pm$ 0.2) & \textbf{89.1} ($\pm$ 0.3) & \underline{91.8} ($\pm$ 0.1) & \textbf{90.6} ($\pm$ 0.1) & \textbf{89.0} ($\pm$ 0.4)  \\

\multicolumn{7}{c}{\cellcolor[HTML]{C0C0C0}SVHN with CNN} \\
FedAvg~\cite{fedavg}                & \underline{92.7} ($\pm$ 0.1) & \underline{90.9} ($\pm$ 0.1) & \underline{89.2} ($\pm$ 0.2) & \underline{92.1} ($\pm$ 0.1) & \underline{89.7} ($\pm$ 0.3) & \underline{88.9} ($\pm$ 0.2)  \\
SignSGD~\cite{signsgd}              & 92.3 ($\pm$ 0.1) & 80.1 ($\pm$ 1.5) & 66.1 ($\pm$ 1.3) & 90.7 ($\pm$ 0.1) & 80.4 ($\pm$ 1.4) & 64.7 ($\pm$ 1.9)  \\
EF-SignSGD~\cite{ef-signsgd}        & 92.6 ($\pm$ 0.2) & 90.7 ($\pm$ 0.1) & 88.3 ($\pm$ 0.1) & 91.8 ($\pm$ 0.1) & 89.2 ($\pm$ 0.2) & 86.8 ($\pm$ 0.4)  \\
Noisy-SignSGD~\cite{noisy-signsgd}  & 92.2 ($\pm$ 0.1) & 90.3 ($\pm$ 0.2) & 88.2 ($\pm$ 0.1) & 90.9 ($\pm$ 0.2) & 88.7 ($\pm$ 0.1) & 86.9 ($\pm$ 0.2)  \\
Stoc-SignSGD~\cite{stoc-signsgd}    & 92.3 ($\pm$ 0.1) & 88.0 ($\pm$ 0.6) & 86.7 ($\pm$ 0.8) & 90.7 ($\pm$ 0.2) & 87.7 ($\pm$ 0.3) & 84.8 ($\pm$ 0.2)  \\
ZeroFL~\cite{zerofl}                & 91.7 ($\pm$ 0.1) & 90.2 ($\pm$ 0.1) & 87.6 ($\pm$ 0.3) & 90.3 ($\pm$ 0.1) & 88.4 ($\pm$ 0.3) & 87.0 ($\pm$ 0.2)  \\
FedBAT                               & \textbf{92.9} ($\pm$ 0.1) & \textbf{91.1} ($\pm$ 0.1) & \textbf{89.3} ($\pm$ 0.2) & \textbf{92.5} ($\pm$ 0.1) & \textbf{90.7} ($\pm$ 0.1) & \textbf{89.2} ($\pm$ 0.2)  \\

\multicolumn{7}{c}{\cellcolor[HTML]{C0C0C0}CIFAR-10 with ResNet-10} \\
FedAvg~\cite{fedavg}                & \textbf{91.5} ($\pm$ 0.1) & \textbf{89.0} ($\pm$ 0.2) & \textbf{83.7} ($\pm$ 0.4) & \textbf{89.3} ($\pm$ 0.1) & \underline{84.6} ($\pm$ 0.2) & \underline{80.9} ($\pm$ 0.5)  \\
SignSGD~\cite{signsgd}              & 88.9 ($\pm$ 0.1) & 87.8 ($\pm$ 0.2) & 76.1 ($\pm$ 1.6) & 87.3 ($\pm$ 0.2) & 82.2 ($\pm$ 0.4) & 76.6 ($\pm$ 0.9)  \\
EF-SignSGD~\cite{ef-signsgd}        & 90.8 ($\pm$ 0.1) & 87.4 ($\pm$ 0.2) & 78.3 ($\pm$ 0.9) & 87.4 ($\pm$ 0.2) & 81.8 ($\pm$ 0.5) & 76.6 ($\pm$ 0.9)  \\
Noisy-SignSGD~\cite{noisy-signsgd}  & 90.1 ($\pm$ 0.2) & 86.2 ($\pm$ 0.1) & 78.3 ($\pm$ 0.7) & 85.5 ($\pm$ 0.2) & 80.2 ($\pm$ 0.3) & 72.7 ($\pm$ 0.4)  \\
Stoc-SignSGD~\cite{stoc-signsgd}    & 88.7 ($\pm$ 0.2) & 86.2 ($\pm$ 0.2) & 77.8 ($\pm$ 0.4) & 85.9 ($\pm$ 0.2) & 80.5 ($\pm$ 0.3) & 74.1 ($\pm$ 1.0)  \\
ZeroFL~\cite{zerofl}                & 89.0 ($\pm$ 0.1) & 86.3 ($\pm$ 0.1) & 79.0 ($\pm$ 0.6) & 85.2 ($\pm$ 0.2) & 78.4 ($\pm$ 0.2) & 73.8 ($\pm$ 0.6)  \\
FedBAT                               & \underline{91.2} ($\pm$ 0.1) & \underline{88.6} ($\pm$ 0.1) & \underline{82.8} ($\pm$ 0.1) & \underline{89.2} ($\pm$ 0.2) & \textbf{84.9} ($\pm$ 0.3) & \textbf{81.0} ($\pm$ 0.6)  \\

\multicolumn{7}{c}{\cellcolor[HTML]{C0C0C0}CIFAR-100 with ResNet-10} \\
FedAvg~\cite{fedavg}                & \textbf{67.7} ($\pm$ 0.2) & \textbf{64.1} ($\pm$ 0.3) & \textbf{54.5} ($\pm$ 0.4) & \textbf{59.2} ($\pm$ 0.3) & \textbf{55.4} ($\pm$ 0.8) & \underline{49.0} ($\pm$ 0.6)  \\
SignSGD~\cite{signsgd}              & 58.9 ($\pm$ 0.6) & 53.9 ($\pm$ 0.2) & 34.3 ($\pm$ 1.5) & 54.2 ($\pm$ 0.3) & 39.3 ($\pm$ 0.4) & 32.5 ($\pm$ 1.7)  \\
EF-SignSGD~\cite{ef-signsgd}        & 65.6 ($\pm$ 0.2) & 59.7 ($\pm$ 0.4) & 50.7 ($\pm$ 0.4) & 53.8 ($\pm$ 0.5) & 46.8 ($\pm$ 0.5) & 40.2 ($\pm$ 0.6)  \\
Noisy-SignSGD~\cite{noisy-signsgd}  & 65.3 ($\pm$ 0.2) & 58.3 ($\pm$ 0.2) & 46.6 ($\pm$ 0.2) & 52.6 ($\pm$ 0.6) & 46.2 ($\pm$ 0.5) & 38.3 ($\pm$ 0.3)  \\
Stoc-SignSGD~\cite{stoc-signsgd}    & 61.1 ($\pm$ 0.4) & 57.8 ($\pm$ 0.4) & 46.2 ($\pm$ 0.7) & 54.2 ($\pm$ 0.2) & 47.2 ($\pm$ 0.3) & 40.1 ($\pm$ 0.6)  \\
ZeroFL~\cite{zerofl}                & 63.7 ($\pm$ 0.2) & 59.9 ($\pm$ 0.5) & 47.7 ($\pm$ 0.8) & 50.5 ($\pm$ 0.5) & 45.6 ($\pm$ 0.4) & 36.1 ($\pm$ 0.5)  \\
FedBAT                               & \underline{66.3} ($\pm$ 0.1) & \underline{63.9} ($\pm$ 0.4) & \underline{53.9} ($\pm$ 0.3) & \underline{58.6} ($\pm$ 0.3) & \underline{54.3} ($\pm$ 0.4) & \textbf{49.2} ($\pm$ 0.6)  \\

\bottomrule[1pt]
\end{tabular}}
\end{table*}

\subsection{Experimental Setup} 
\textbf{Datasets and Models.} 
In this section, we evaluate FedBAT using four widely recognized datasets: FMNIST~\cite{fmnist}, SVHN~\cite{svhn}, CIFAR-10 and CIFAR-100~\cite{cifar10}. To showcase the versatility of FedBAT, we also assess its performance across different model architectures. Specifically, we employ a CNN with four convolution layers and one fully connected layer for FMNIST and SVHN, and ResNet-10~\cite{resnet} for CIFAR-10 and CIFAR-100. Detailed model architectures are available in Appendix~\ref{sec:a1}. 

\textbf{Data Partitioning.} 
We consider both cases of IID and Non-IID data distribution, referring to the data partitioning benchmark of FL~\cite{noniid-benchmark}. Under IID partitioning, an equal quantity of data is randomly sampled for each client. The Non-IID scenario further encompasses two distinct label distributions, termed Non-IID-1 and Non-IID-2. 
In Non-IID-1, the proportion of the same label among clients follows the Dirichlet distribution~\cite{dirichlet}, while in Non-IID-2, each client only contains data of partial labels. For CIFAR-100, we set the Dirichlet parameter to 0.1 in Non-IID-1 and assign 10 random labels to each client in Non-IID-2. For the other datasets, we set the Dirichlet parameter to 0.3 in Non-IID-1 and assign 3 random labels to each client in Non-IID-2. 

\textbf{Baseline Methods.} 
All experiments are conducted on Flower~\cite{flower}, an open-source training platform for FL. FedAvg~\cite{fedavg} is adopted as the backbone training algorithm. We compare FedBAT with the binarization methods discussed in Section~\ref{sec:intro}, including SignSGD~\cite{signsgd}, EF-SignSGD~\cite{ef-signsgd}, Noisy-SignSGD~\cite{noisy-signsgd} and Stoc-SignSGD~\cite{stoc-signsgd}. We also compare FedBAT with ZeroFL \cite{zerofl}, a method that exploits local sparsity to compress communications. To ensure similar traffic volume to the binarization methods, we set the sparsity ratio of ZeroFL to 97\%. Details about the baselines are provided in Appendix~\ref{sec:a2}. 

\textbf{Hyperparameters.} 
The number of clients is set to 30 and 100, respectively. 10 clients will participate in every round. The local epoch is set to 10 and the batch size is set to 64. SGD~\cite{sgd} is used as the local optimizer. The learning rate is tuned from (1.0, 0.1, 0.01) and set to 0.1. The number of rounds are set to 100 for CNN and 200 for ResNet-10. 
For the baselines, each hyperparameter is carefully tuned among (1.0, 0.1, 0.01, 0.001), including the step size and the coefficient of noise. Detailed tuning process and results are provided in Appendix~\ref{sec:a2}. In FedBAT, the coefficient $\rho$ is set to 6 and the warm-up ratio $\phi$ is set to 0.5 by default. Each experiment is run five times on Nvidia 3090 GPUs with Intel Xeon E5-2673 CPUs. Average results and the standard deviation are reported.  

\subsection{Overall Performance}\label{sec:overall}
In this subsection, we compare the performance of FedBAT and the baselines by the test accuracy and the convergence speed. All numerical results are reported in Table~\ref{tb:performance}. The convergence curves on CIFAR-100 with 100 clients are shown in Figure~\ref{fig:converge}. The convergence curves on other datasets are provided in Appendix~\ref{sec:b3}. In addition, experimental results of more clients and more related baselines are also available in Appendix~\ref{sec:b}.

As shown in Table~\ref{tb:performance}, compared with SignSGD, other binarization baselines can indeed improve the test accuracy in most cases. However, in a few cases, such as CIFAR-10 under the Non-IID-1 data distribution, their accuracy can be even worse than SignSGD, which reveals the limitations of existing binarization methods. For ZeroFL, we observe a generally lower accuracy compared to the binarization methods. This discrepancy can be attributed to the higher sparsity ratio, which hinders the update of most parameters. In addition, all baselines suffer significant accuracy loss compared to FedAvg, particularly in scenarios involving high degrees of Non-IID data distribution. On the contrary, FedBAT can generally achieve comparable or even higher accuracy than FedAvg,  irrespective of data distribution or the number of clients.  In terms of convergence speed, as shown in the Figure~\ref{fig:converge}, FedBAT can consistently outperform all binarization baselines and approach that of FedAvg. 

\subsection{Ablation Studies}
In this subsection, we conduct ablation studies to assess the feasibility and superiority of FedBAT's design. Specifically, we adjust the coefficient of the step size $\rho$ and the local warm-up ratio $\phi$ in the context of the Non-IID-2 data distribution with 100 clients. We show in the subsequent experiments that FedBAT also outperforms the baseline methods in terms of hyperparameter tuning. 

\begin{table}[t]
\renewcommand{\arraystretch}{1.0}
\centering
\caption{The test accuracy of FedBAT with varying $\rho$.}\label{tb:ablation-k}
\scalebox{0.75}{
\begin{tabular}{ccccccc}
\toprule[1pt]
            & $\rho=0$  & $\rho=2$  & $\rho=4$  & $\rho=6$  & $\rho=8$  & $\rho=10$   \\ \midrule
FMNIST      & 87.9   & 88.2   & \underline{88.9}   & \textbf{89.0}   & 88.7   & \underline{88.9}     \\
SVHN        & 88.5   & 89.0   & \underline{89.2}   & \underline{89.2}   & \textbf{89.5}   & 88.7     \\
CIFAR-10    & 78.3   & 79.0   & 80.4   & \textbf{81.0}   & \underline{80.6}   & 80.1     \\
CIFAR-100   & 41.2   & 48.0   & 48.7   & \textbf{49.2}   & \underline{48.8}   & 48.7     \\
\bottomrule[1pt]
\end{tabular}}
\end{table}

We first explore the FedBAT variant with $\rho=0$, where the step size $\alpha$ remains fixed at its initial value $\alpha'$ without undergoing any optimization. Comparing Table~\ref{tb:performance} and Table~\ref{tb:ablation-k}, it can be found that FedBAT ($\rho=0$) still achieves better accuracy than all binarization baselines, which proves the superiority of binarizing model updates during local training. Nevertheless, the accuracy of FedBAT ($\rho=0$) remains inferior to FedAvg, which underscores the need for an adaptive step size. Therefore, we further tune $\rho$ among $\{2, 4, 6, 8, 10\}$ to verify the effect and robustness of learning the step size $\alpha$. 
As illustrated in Table~\ref{tb:ablation-k}, setting the parameter $\rho$ to 2 enhances the accuracy of FedBAT, aligning it more closely with FedAvg. The gradual increase in the value of $\rho$ leads to a slight continuous improvement in the accuracy of FedBAT until $\rho$ reaches 10. It is noteworthy that for $\rho$ values of 4, 6, and 8, the accuracy difference in FedBAT is comparatively minimal. This indicates a broad optimal range for the hyperparameter $\rho$, highlighting the robustness of FedBAT to variations in $\rho$.

\begin{table}[t]
\renewcommand{\arraystretch}{1.0}
\centering
\caption{The test accuracy of FedBAT with varying $\phi$.}\label{tb:ablation-r}
\scalebox{0.75}{
\begin{tabular}{cccccc}
\toprule[1pt]
            & $\phi=0.1$   & $\phi=0.3$   & $\phi=0.5$   & $\phi=0.7$   & $\phi=0.9$     \\ \midrule
FMNIST      & 88.6      & \underline{88.7}      & \textbf{89.0}      & \textbf{89.0}      & 88.5 \\
SVHN        & 88.4      & \underline{88.8}      & \textbf{89.2}      & 88.5      & 88.6 \\
CIFAR-10    & 75.9      & 79.9      & \textbf{81.0}      & \underline{80.7}      & 80.3 \\
CIFAR-100   & 42.9      & 48.6      & \textbf{49.2}      & \underline{48.9}      & 48.3 \\
\bottomrule[1pt]
\end{tabular}}
\end{table}
Another hyperparameter of FedBAT is the local warm-up ratio $\phi$, which balances the trade-off between the binarization of model updates and their initialization. Here, we tune $\phi$ among $\{0.1, 0.3, 0.5, 0.7, 0.9\}$. As shown in the Table~\ref{tb:ablation-r}, the optimal accuracy for FedBAT is always achieved when $\phi$ is set to 0.5, indicating an equilibrium between the significance of initializing and binarizing model updates. For $\phi$ values of 0.3 or 0.7, there is a marginal reduction in accuracy, although the variance in accuracy is minimal. Notably, further deviation by increasing $\phi$ to 0.9 or decreasing it to 0.1 leads to more decline in the accuracy.

Following the above ablation studies, we summarize the essential findings regarding hyperparameter tuning about FedBAT. It is advisable to set the local warm-up ratio $\phi$ to 0.5 as a default configuration without necessitating hyperparameter tuning. The default setting for $\rho$ in FedBAT is recommended to be 6, with an option to tune it within the interval of [4,8] for marginal accuracy enhancements.

\section{Related Work}\label{sec:related}
\subsection{Communication-Efficient Federated Learning} 
Existing methods reduce communication costs in FL from two aspects: model compression and gradient compression. The proposed FedBAT belongs to the latter paradigm.

Frequently transferring large models between the server and clients is a significant burden, especially for clients with limited communication bandwidth. Therefore, researchers have turned to model compression techniques. \citet{bifl} train and communicate a binary neural network in FL. Different from their motivation, our focus lies in learning binary model updates rather than binary weights. \cite{feddropout,ada-feddropout} enable clients to train randomly selected sub-models of a larger server model. \citet{fedpara} use matrix factorization to reduce the size of the model that needs to be transferred. \cite{fedmask,fedpm} transfer a pruned model for efficient communication. Specifically, FedPM~\cite{fedpm} trains and communicates only a binary mask for each model parameter, while keeping the model parameters at their randomly initialized values. It is worth noting that FedPM has certain similarities with FedBAT. They both keep the base model parameters fixed during local training. FedPM trains binary masks to prune the base model, while FedBAT trains binary model updates with respect to the base model. The model trained by FedPM enjoys the advantage of sparsity, however, the randomly initialized model parameters do not undergo any updates. Due to the lack of effective training on randomly initialized parameters, FedPM may not be able to achieve a satisfactory accuracy as FedAvg~\cite{hidenseek}.

Apart from the model compression, another way to reduce communication costs is gradient compression. It is generally achieved by pruning~\cite{zerofl} or quantization (including binarization)~\cite{fedpaq,signsgd} on the model updates. To enhance efficiency, recent quantization methods~\cite{adaquantfl,feddq,dadaquant} try to adjust the quantization bitwidth used in different rounds adaptively. However, they suffer from the same post-training manner as existing binarization methods. We posit that the concept of FedBAT can be seamlessly applied to federated quantization.

\subsection{Binarization} 
\vspace{-6pt}
BinaryConnect~\cite{binary-connect} is a pioneering work on learning binary weights. It binarizes weights into $\{-1, +1\}$ to calculate the output loss. During backpropagation, the full precision weights are optimized by STE~\cite{ste}. However, BinaryConnect does not involve learning the step size. Further, BWN and XNOR-Net are proposed to calculate a step size by minimizing the binarized errors~\cite{xnornet}. To be specific, the step size is set to the average of absolute weight values. Later, \citet{lab} propose a proximal Newton algorithm with diagonal Hessian approximation that directly minimizes the loss with respect to the binarized weights and the step size. However, it requires the second-order gradient information, usually not available in SGD. In addition, considering that binarization is essentially a form of 1-bit quantization, many quantization-aware training methods are also suitable for binarization. For example, PACT~\cite{pact} and LSQ~\cite{lsq} achieve end-to-end optimization by designing reasonable gradient for the step size. 

In this paper, we discovered the post-training manner of model updates compression in FL. We seek to solve this problem by directly learning binary model updates during local training. However, the above methods are designed to learn binarized or quantized weights by a long period of centralized training. It is difficult for them to learn accurate binary model updates during a short local training period. Therefore, after designing the gradient of the step size, we also introduced a temperature $\rho$ to regulate the pace of its optimization, along with the implementation of warm-up training for improved initialization.

\vspace{-6pt}
\section{Conclusion}
\vspace{-6pt}
We analyzed the challenges faced by existing binarization methods when applied in the context of FL. The analysis encourages us to leverage the local training process to learn binary model updates, instead of binarizing them after training. Therefore, we propose FedBAT, a federated training framework designed with a focus on binarization awareness. FedBAT has undergone comprehensive theoretical examination and experimental validation. It is able to exceed the binarization baselines in terms of the test accuracy
and convergence speed, and is comparable to FedAvg. 

An inherent limitation of FedBAT resides in the requirement for the client to retain two distinct copies of model parameters: one for the initialization model and another for model updates. Although only one copy of parameters shall be trained, this slightly increases the memory overhead of the local training process in FedBAT. However, by compressing the initialization model transmitted from the server before communication, the memory overhead in FedBAT can be alleviated, as well as the downlink communication can be further compressed. We leave this as future work.

\section*{Acknowledgements}
This work is supported in part by National Natural Science Foundation of China under grants 62376103, 62302184, 62206102, Science and Technology Support Program of Hubei Province under grant 2022BAA046, and Ant Group through CCF-Ant Research Fund.
\section*{Impact Statement}
In this paper, we propose FedBAT to improve the communication efficiency of federated learning. It is our contention that FedBAT does not result in any adverse social impact. Instead, FedBAT enhances user privacy protection to a certain extent as the model updates uploaded by the clients are binarized and noise is introduced in the process.

\bibliography{icml2024}
\bibliographystyle{icml2024}
\newpage
\appendix\label{sec:appendix}
\onecolumn

\section{Detailed Experimental Settings}\label{sec:a}
\subsection{Model Architectures}\label{sec:a1}
In this paper, we employ a CNN for FMNIST and SVHN, ResNet-10 for CIFAR-10 and CIFAR-100. The detailed model architectures are shown in Table~\ref{tb:models}. The BasicBlock used by ResNet is the same as defined in~\citep{resnet}. Batch normalization (BN)~\citep{bn} is used to ensure stable training and ReLU~\cite{relu} is employed as the activation function.

\begin{table*}[h]
\renewcommand{\arraystretch}{1.0}
\centering
\caption{Model architectures of the CNN, ResNet-10.}\label{tb:models}
\scalebox{0.8}{
\begin{tabular}{cccc}
\toprule[1pt]
\makebox[0.18\textwidth][c]{CNN (FMNIST)} & \makebox[0.18\textwidth][c]{CNN (SVHN)} & \makebox[0.18\textwidth][c]{ResNet-10 (CIFAR-10)} & \makebox[0.18\textwidth][c]{ResNet-10 (CIFAR-100)} \\ \toprule[1pt]
Convd2d(1,32,3)        & Convd2d(3,32,3)       & Convd2d(3,32,3)    & Convd2d(3,32,3)   \\ \midrule
Convd2d(32,64,3)       & Convd2d(32,64,3)      & BasicBlock(32)     & BasicBlock(32)    \\ \midrule
Convd2d(64,128,3)      & Convd2d(64,128,3)     & BasicBlock(64)     & BasicBlock(64)    \\ \midrule
Convd2d(128,256,3)     & Convd2d(128,256,3)    & BasicBlock(128)    & BasicBlock(128)   \\ \midrule
Linear(256,10)         & Linear(1024,10)       & BasicBlock(256)    & BasicBlock(256)   \\ \midrule
                       &                       & Linear(256,10)     & Linear(256,100)    \\
\bottomrule[1pt]
\end{tabular}}
\end{table*}

\subsection{Baseline Methods}\label{sec:a2}
In our experiments, we compare FedBAT with five baselines, including SignSGD~\citep{signsgd}, EF-SignSGD~\citep{ef-signsgd}, Noisy-SignSGD~\citep{noisy-signsgd}, Stoc-SignSGD~\citep{stoc-signsgd} and ZeroFL~\cite{zerofl}. Here, we introduce each baseline and the hyperparameters involved in detail. We declare that all hyperparameters are tuned in $\{1.0, 0.1, 0.01, 0.001\}$ for each dataset. 
The hyperparameter of SignSGD is the step size $\alpha$, which is tuned and set to 0.001 for all datasets. In EF-SignSGD, there is no hyperparameter to tune. Notably, EF-SignSGD adds the binarization errors from the previous round onto the current model updates before conducting another binarization. Furthermore, it sets the step size $\alpha$ to $\nicefrac{\Vert \boldsymbol{m}\Vert _1}{d}$ for a more precise binarization of $\boldsymbol{m}\in\mathbb{R}^d$. 
Noisy-SignSGD adds Gaussian noise $\xi\sim N(0,\sigma^2)$ to the model update and then performs binarization, where $\sigma$ is an adjustable standard deviation. We set the step size $\alpha$ and the standard deviation $\sigma$ to 0.01 and 0.01 for Noisy-SignSGD. 
Stoc-SignSGD adds uniform noise $\xi\sim U(-\Vert \boldsymbol{m}\Vert ,\Vert \boldsymbol{m}\Vert )$ to a model update $\boldsymbol{m}$ before binarization. In this way, each element $\boldsymbol{m}_i$ will be binarized into +1 with probability ($\nicefrac{1}{2}+\nicefrac{\boldsymbol{m}_i}{2\Vert \boldsymbol{m}\Vert }$) and into -1 with probability ($\nicefrac{1}{2}-\nicefrac{\boldsymbol{m}_i}{2\Vert \boldsymbol{m}\Vert }$). However, we observe that the value of $\Vert \boldsymbol{m}\Vert $ is so large that the above two probabilities are both close to 0.5. Therefore, in our experiments, we replaced $\Vert \boldsymbol{m}\Vert $ with $\Vert \boldsymbol{m}\Vert _\infty$ for a better binarization. Besides, the step size of Stoc-SignSGD is tuned and set to 0.01. ZeroFL performs SWAT~\cite{swat} in local training and prune the model updates after training. The sparsity ratio is set to 97\% to ensure a communication compression ratio similar to the binarization methods.

\section{Additional Experiment Results}\label{sec:b}
\subsection{Additional Baselines}\label{sec:b1}
In this subsection, we test additional baselines on CIFAR-10, including FedPAQ~\cite{fedpaq} and BiFL~\cite{bifl}. FedPAQ quantizes model updates after local training. Table~\ref{tb:fedpaq} shows that FedPAQ achieve comparable accuracy to FedBAT with a communication costs ranging from 3 to 4 bits per parameter (bpp). BiFL trains a binary neural network (BNN) during local training. There are two ways to upload local weights in BiFL, one is to communicate full-precision weights (BiFL-Full), and the other is to communicate binarized weights (BiFL). BiFL-Full achieves higher accuracy as communication is uncompressed. However, BiFL directly binarizes model weights rather than binarizing model updates, resulting in much lower accuracy compared to SignSGD.

\begin{table*}[h]
\renewcommand{\arraystretch}{1.0}
\centering
\caption{The test accuracy of additional baselines (including FedPAQ and BiFL) on CIFAR-10 with 100 clients.}\label{tb:fedpaq}
\scalebox{0.75}{
\begin{tabular}{lccccccc}
\toprule[1pt]

& \makebox[0.13\textwidth][c]{FedAvg}  & \makebox[0.13\textwidth][c]{SignSGD} & \makebox[0.13\textwidth][c]{FedPAQ [3-bit]} & \makebox[0.13\textwidth][c]{FedPAQ [4-bit]}  & \makebox[0.13\textwidth][c]{BiFL} & \makebox[0.13\textwidth][c]{BiFL-Full} & \makebox[0.13\textwidth][c]{FedBAT} \\ 
\hline

IID             & 89.3 & 87.3 & 88.8 & 89.1 & 61.8 & 86.8 & 89.2   \\
Non-IID-1       & 84.6 & 82.2 & 83.9 & 84.5 & 37.4 & 84.3 & 84.9   \\
Non-IID-2       & 80.9 & 76.6 & 80.5 & 80.8 & 25.1 & 79.6 & 81.0   \\

\bottomrule[1pt]
\end{tabular}}
\end{table*}

\subsection{More Clients}\label{sec:b2}
In Section~\ref{sec:exp}, the number of clients is set to 30 and 100. Now, we expand the number of clients to 200. As shown in Table~\ref{tb:200_clients}, despite this increase, FedBAT consistently achieves accuracy comparable to FedAvg and notably surpasses SignSGD. 
\vspace{-14pt}
\begin{table*}[ht]
\renewcommand{\arraystretch}{1.0}
\centering
\caption{The test accuracy of FedAvg, SignSGD and FedBAT with 200 clients.}\label{tb:200_clients}
\scalebox{0.75}{
\begin{tabular}{l cccccc cccccc}
\toprule[1pt]

& \multicolumn{3}{c}{FMNIST with CNN} 
& \multicolumn{3}{c}{SVHN with CNN}
& \multicolumn{3}{c}{CIFAR-10 with ResNet-10} 
& \multicolumn{3}{c}{CIFAR-100 with ResNet-10}\\
\cmidrule(r){2-4} \cmidrule(r){5-7} \cmidrule(r){8-10} \cmidrule(r){11-13}

& \makebox[0.075\textwidth][c]{IID} & \makebox[0.075\textwidth][c]{Non-IID-1} & \makebox[0.075\textwidth][c]{Non-IID-2}
& \makebox[0.075\textwidth][c]{IID} & \makebox[0.075\textwidth][c]{Non-IID-1} & \makebox[0.075\textwidth][c]{Non-IID-2}
& \makebox[0.075\textwidth][c]{IID} & \makebox[0.075\textwidth][c]{Non-IID-1} & \makebox[0.075\textwidth][c]{Non-IID-2}
& \makebox[0.075\textwidth][c]{IID} & \makebox[0.075\textwidth][c]{Non-IID-1} & \makebox[0.075\textwidth][c]{Non-IID-2}\\ 

FedAvg      & 91.6  & 90.2 & 88.4 & 91.4 & 88.6 & 87.8  & 84.9  & 81.0 & 78.2 & 49.7 & 48.8 & 42.9   \\
SignSGD     & 89.5  & 87.5 & 80.2 & 89.6 & 79.6 & 67.6  & 84.2  & 77.8 & 71.0 & 46.1 & 34.2 & 30.8   \\
FedBAT      & 91.4  & 89.9 & 88.1 & 91.4 & 89.2 & 87.8  & 85.0  & 80.8 & 77.8 & 49.1 & 48.4 & 43.2   \\

\bottomrule[1pt]
\end{tabular}}
\end{table*}
\vspace{-14pt}

\subsection{Additional Convergence Curves}\label{sec:b3}
\begin{figure*}[!ht]
\centering
\subfigure[FMNIST, IID]{
\begin{minipage}[h]{0.31\textwidth}
\centering
\includegraphics[scale=0.36]{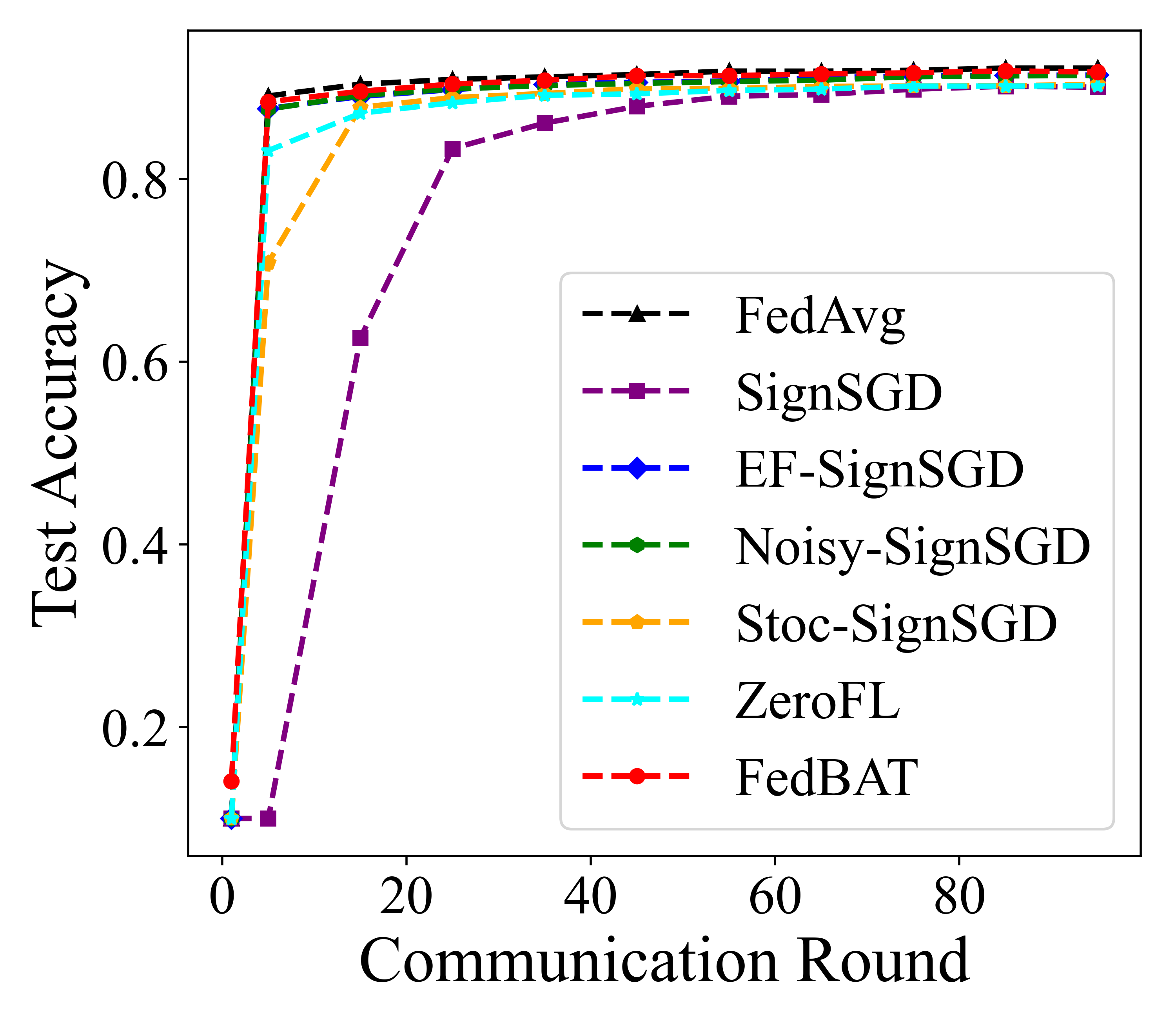}
\end{minipage}
}
\subfigure[FMNIST, Non-IID-1]{
\begin{minipage}[h]{0.31\textwidth}
\centering
\includegraphics[scale=0.36]{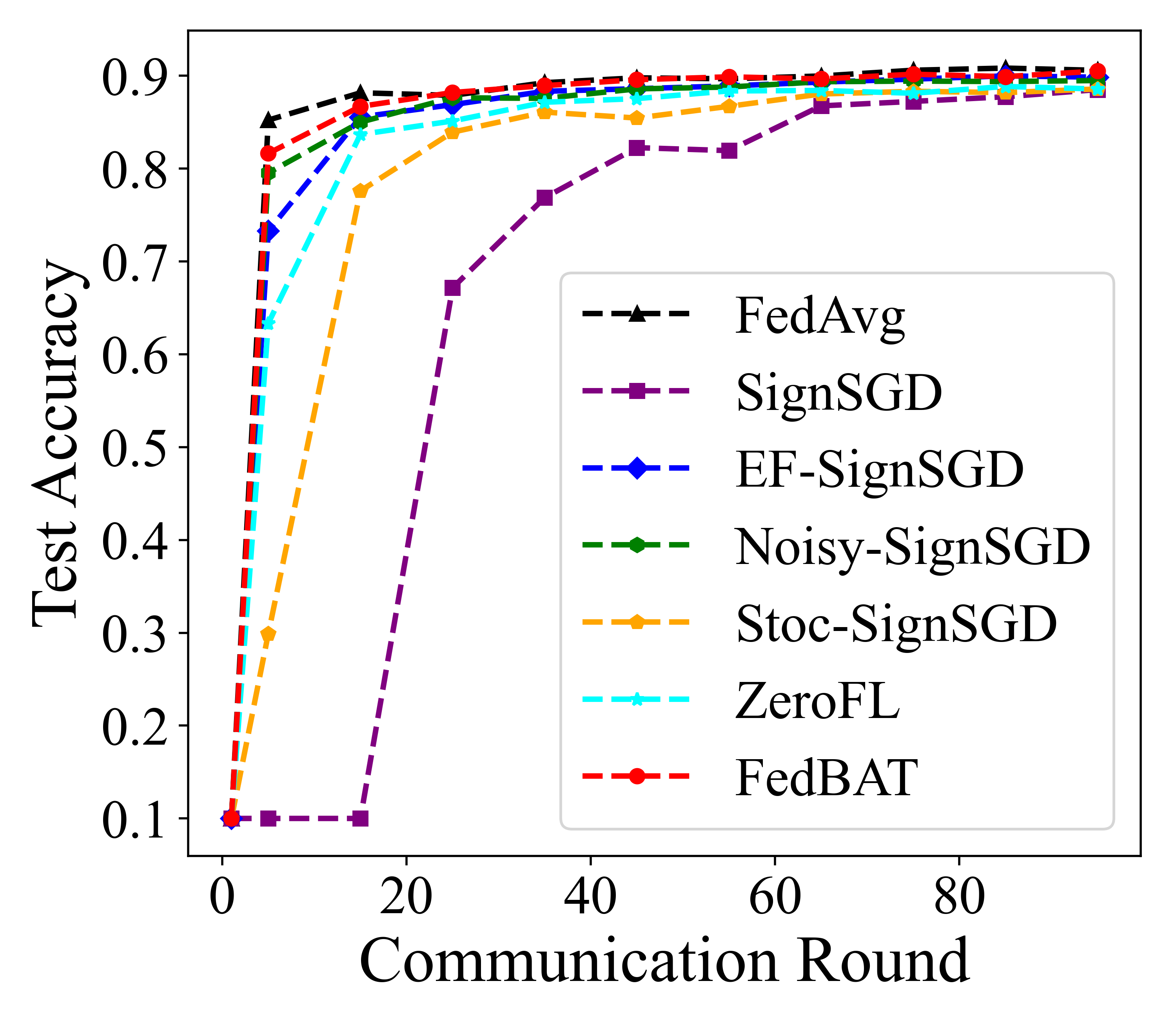}
\end{minipage}
}
\subfigure[FMNIST, Non-IID-2]{
\begin{minipage}[h]{0.31\textwidth}
\centering
\includegraphics[scale=0.36]{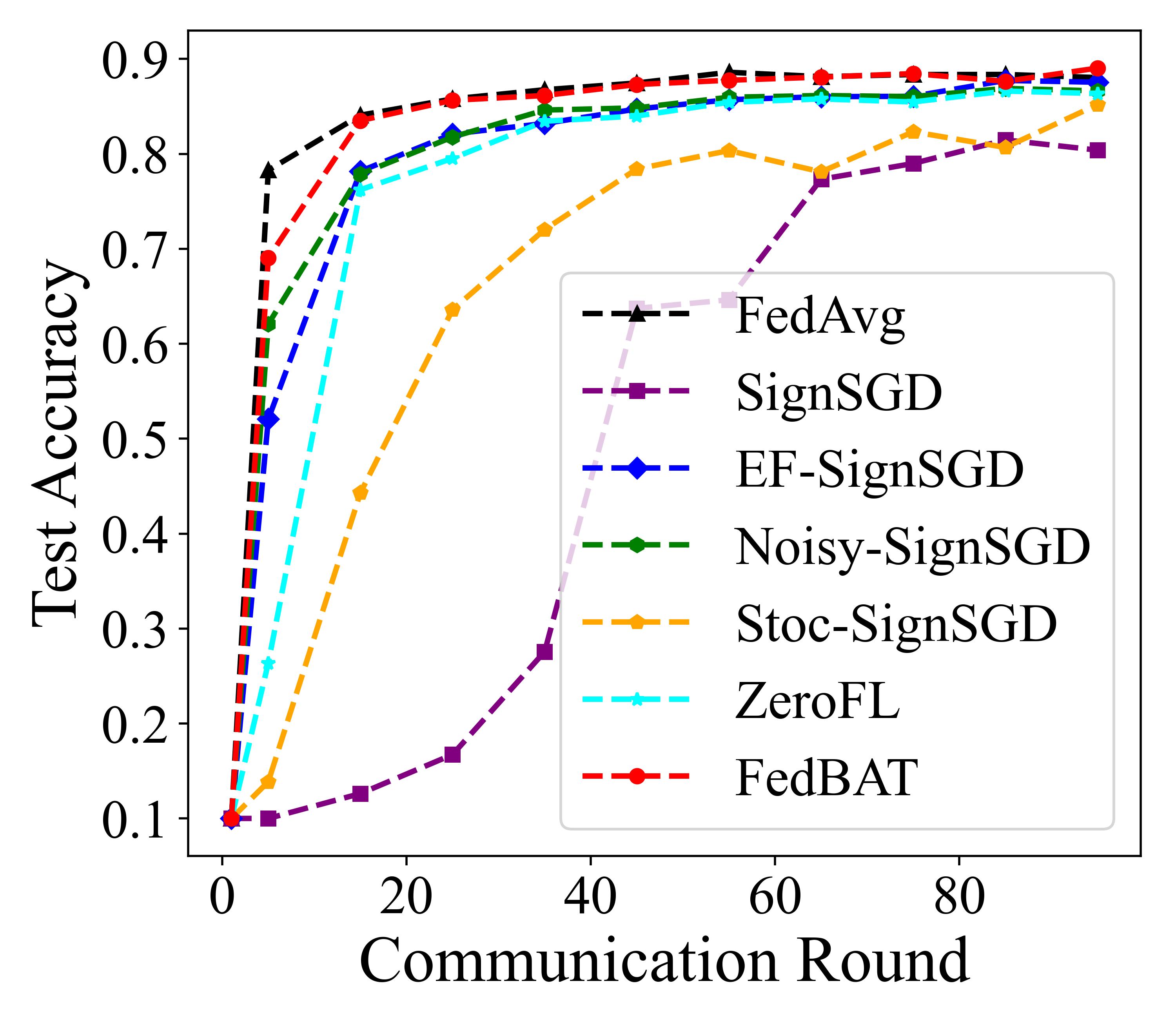}
\end{minipage}
}

\subfigure[SVHN, IID]{
\begin{minipage}[h]{0.31\textwidth}
\centering
\includegraphics[scale=0.36]{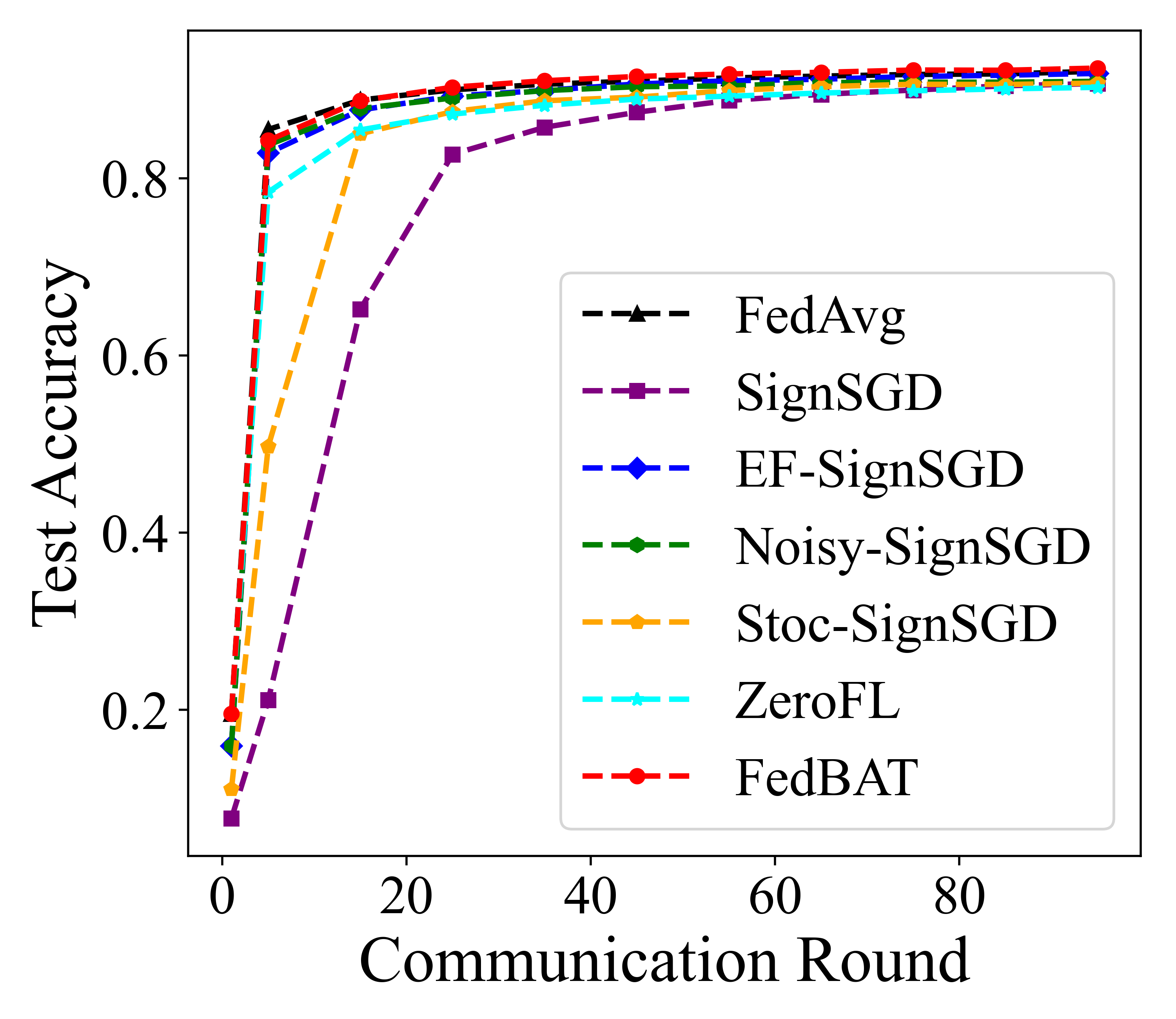}
\end{minipage}
}
\subfigure[SVHN, Non-IID-1]{
\begin{minipage}[h]{0.31\textwidth}
\centering
\includegraphics[scale=0.36]{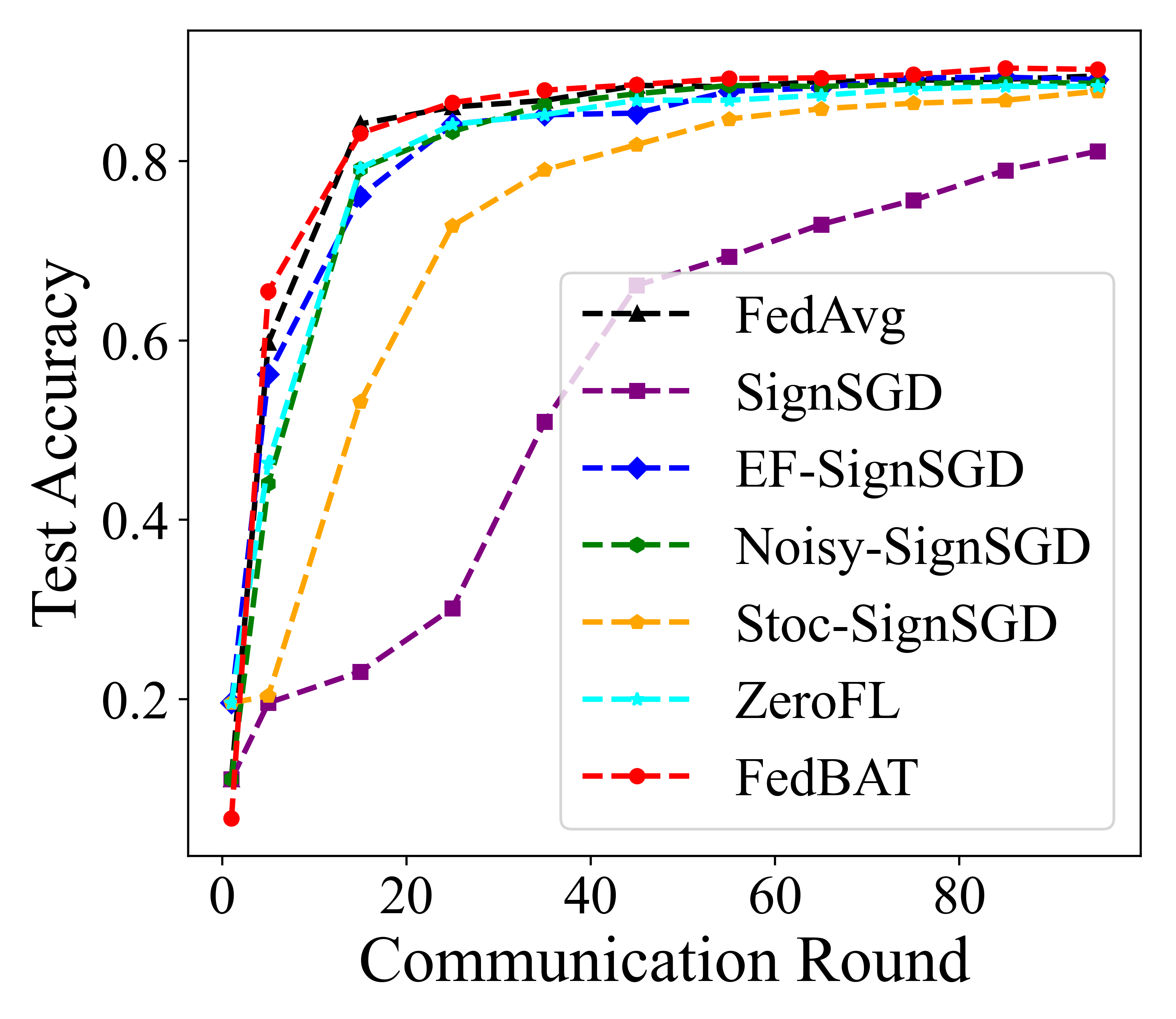}
\end{minipage}
}
\subfigure[SVHN, Non-IID-2]{
\begin{minipage}[h]{0.31\textwidth}
\centering
\includegraphics[scale=0.36]{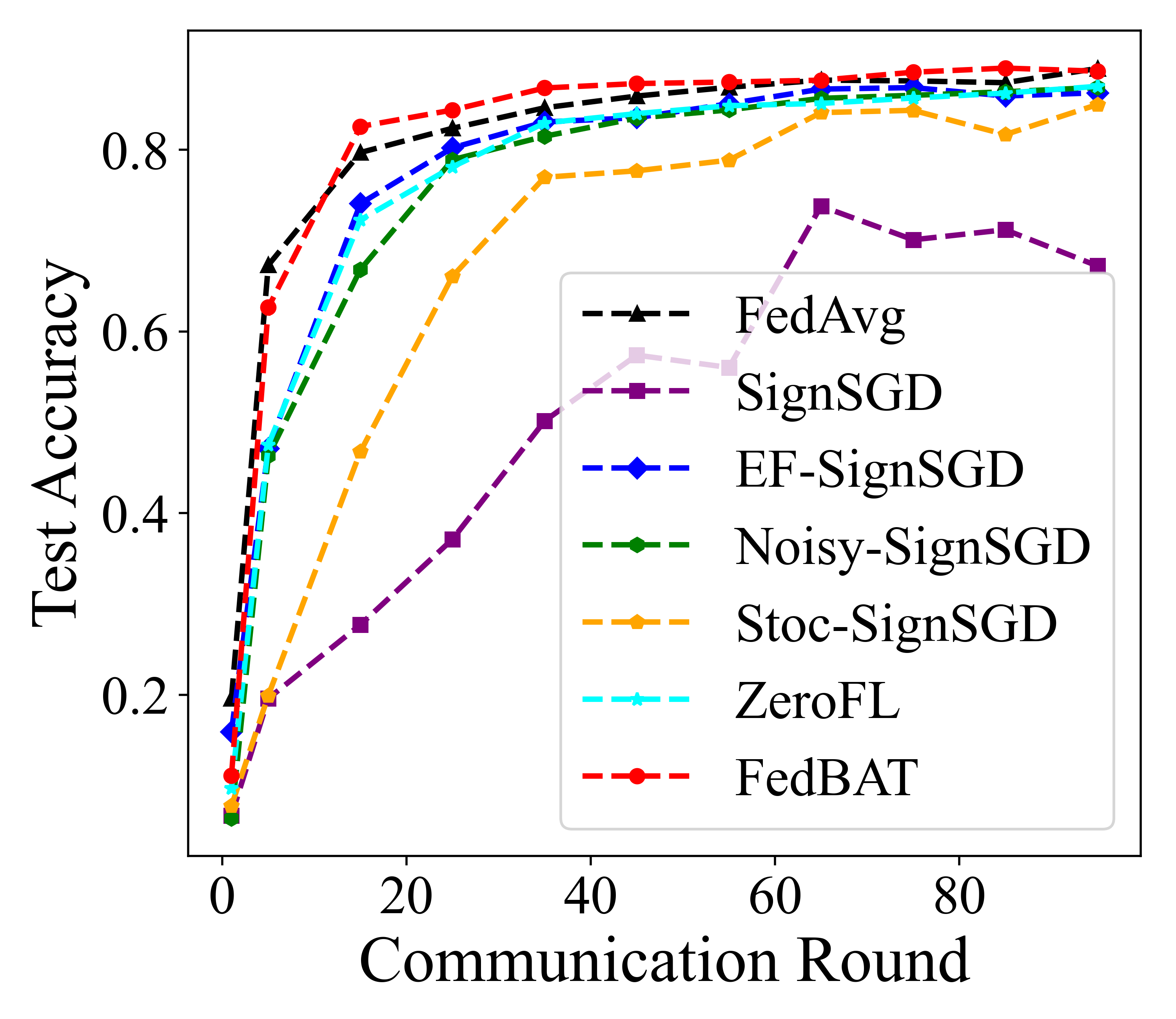}
\end{minipage}
}
\subfigure[CIFAR-10, IID]{
\begin{minipage}[h]{0.31\textwidth}
\centering
\includegraphics[scale=0.36]{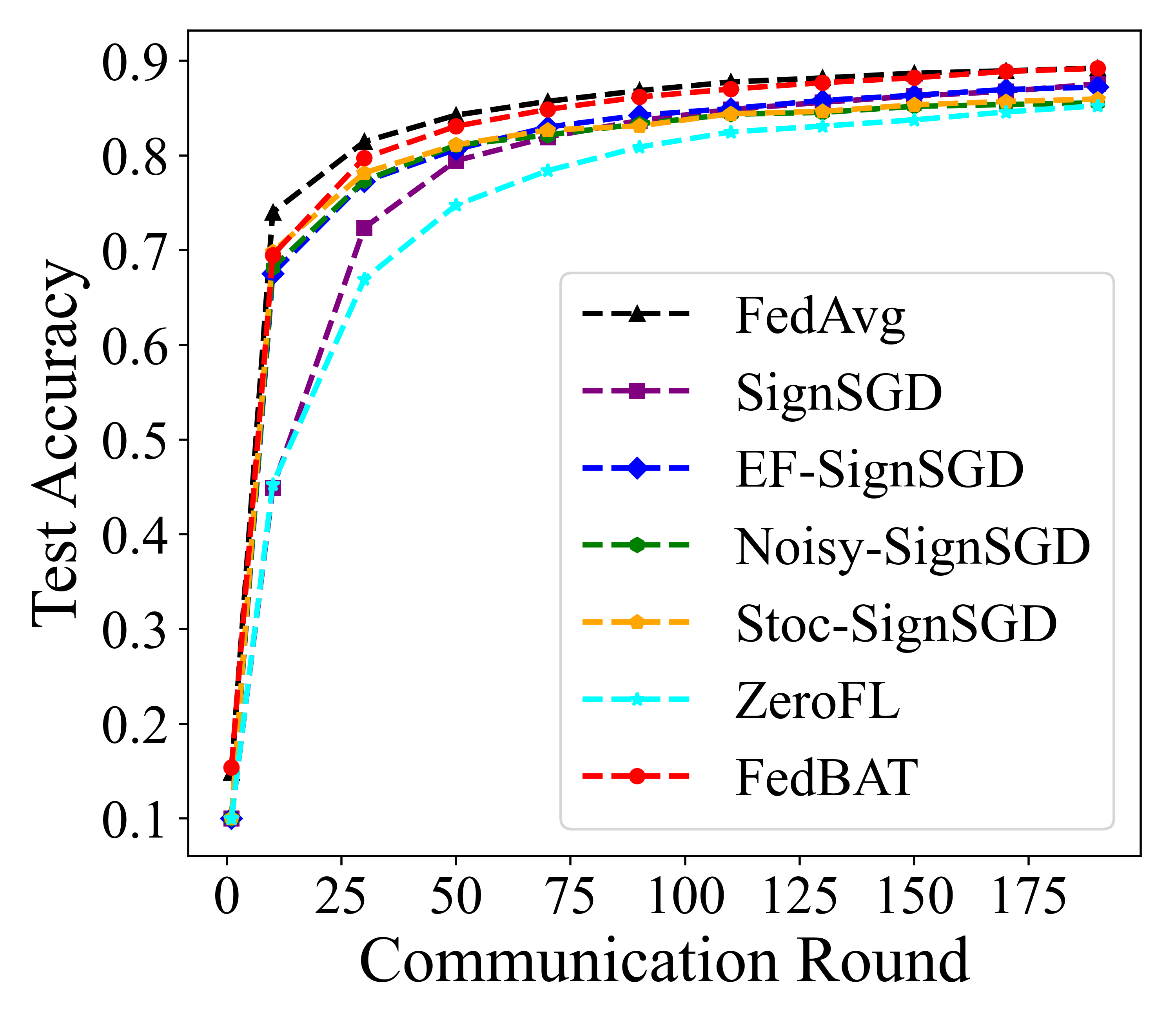}
\end{minipage}
}
\subfigure[CIFAR-10, Non-IID-1]{
\begin{minipage}[h]{0.31\textwidth}
\centering
\includegraphics[scale=0.36]{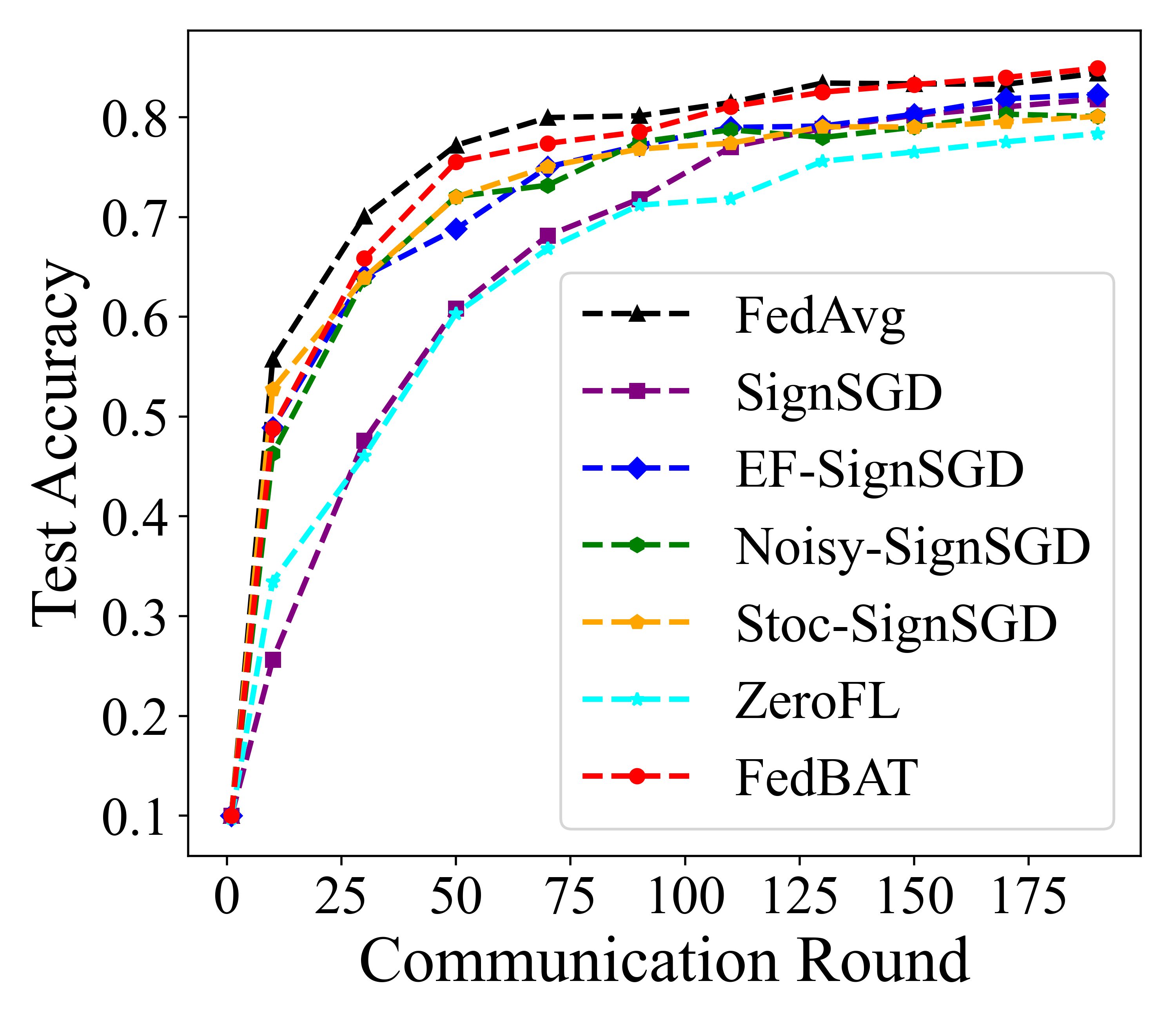}
\end{minipage}
}
\subfigure[CIFAR-10, Non-IID-2]{
\begin{minipage}[h]{0.31\textwidth}
\centering
\includegraphics[scale=0.36]{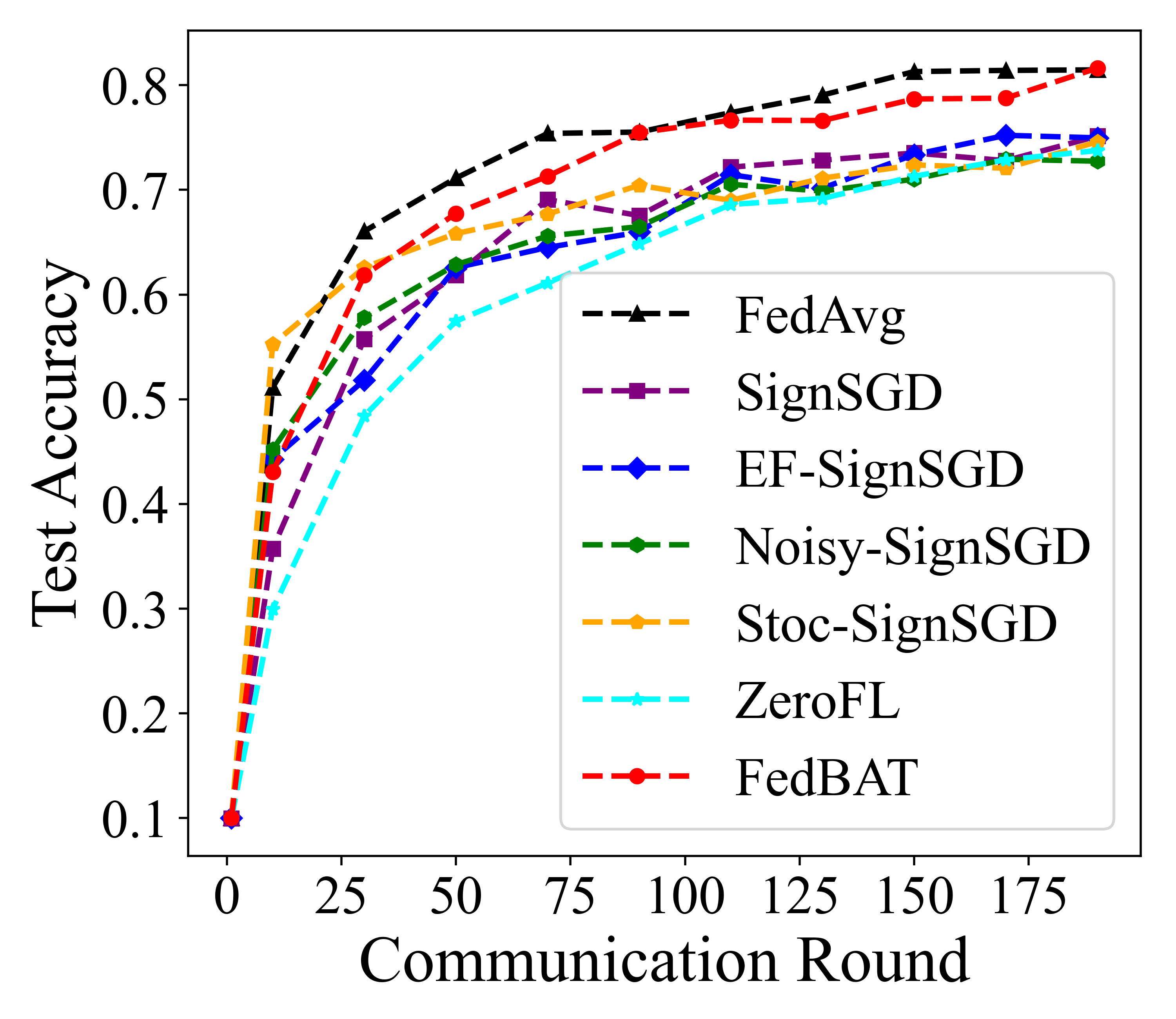}
\end{minipage}
}
\caption{Convergence curves of FedBAT and baselines on FMNIST, SVHN and CIFAR-10 with 100 clients.}\label{fig:converge_appendix}
\end{figure*}

Due to limited space, we illustrate the convergence curves of various methods for the FMNIST, SVHN and CIFAR-10 datasets in Figure~\ref{fig:converge_appendix}. The convergence patterns of the methods closely resemble those observed in Figure~\ref{fig:converge}.
\vspace{-14pt}

\clearpage

\section{Proof of Theorem \ref{thm:full}}\label{sec:proof_full} 
\def \m{\boldsymbol{m}}
\def \hm{\hat{\boldsymbol{m}}}
\def \e{\mathbf{e}}

\def \E{\mathbb{E}}
\def \S{\mathbb{S}}
\def \v{\mathbf{v}}
\def \w{\mathbf{w}}
\def \bv{\bar{\mathbf{v}}}
\def \bw{\bar{\mathbf{w}}}
\def \gt{\mathbf{g}_t}
\def \bvt{\bar{\mathbf{v}}_t}
\def \bwt{\bar{\mathbf{w}}_t}
\def \bgt{\bar{\mathbf{g}}_t}
\def \x{\mathbf{x}}

\def \et{\eta _t}
\def \gmt{\gamma _t}
\def \ie{\mathcal{I}_\tau}
\def \tk{_{t}^{k}}
\def \ttk{_{t+1}^{k}}
\def \sumk{\sum_{k=1}^N}
\def \sumkp{\sum_{k=1}^N p_k}

In this section, we analyze FedBAT in the setting of full device participation. The theoretical analysis in this paper is rooted in the findings about FedAvg presented in~\cite{fedavg_convergence}.

\subsection{Additional Notation}\label{sec:c1}
Let $\w_t^k$ be the model parameters maintained in the $k$-th device at the $t$-th step. Let $\ie$ be the set of global synchronization steps, i.e., $\ie = \{n\tau | n = 1, 2, ...\}$. If $t+1\in\ie$, i.e., the time step to communication, FedBAT activates all devices. Then the optimization of FedBAT can be described as

\begin{equation}
\begin{aligned}
    \v\ttk = \w\tk - \et \nabla F_k(\x\tk, \xi\tk)
\end{aligned}
\end{equation}
\begin{equation}
\begin{aligned}
    \x_{t}^{k} = \mathcal{S}_m(\w\tk)
\end{aligned}
\end{equation}
\begin{equation}
\begin{aligned}
\w\ttk = \left\{
\begin{array}{ll}
    \v\ttk          & \text{if} \ t+1 \notin \ie,\\ 
    \sumkp \mathcal{S}_m(\v\ttk)   & \text{if} \ t+1 \in    \ie. \\
\end{array}
\right. \\
\end{aligned}
\end{equation}

Here, the variable $\v\ttk$ is introduced to represent the immediate result of one step SGD update from $\w\tk$. We interpret $\w\ttk$ as the parameters obtained after communication steps (if possible). 
Also, an additional variable $\x\tk$ is introduced to represent the result of performing binarization on model updates. 

In our analysis, we define two virtual sequences $\bvt=\sumkp \v\tk$ and $\bwt=\sumkp \w\tk$. $\bv_{t+1}$ results from an single step of SGD from $\bwt$. When $t+1 \notin \ie$, both are inaccessible. When $t+1 \in \ie$, we can only fetch $\bw_{t+1}$. For convenience, we define $\bgt = \sumkp \nabla F_k(\x\tk)$ and $\gt = \sumkp \nabla F_k(\x\tk, \xi\tk)$. Therefore, $\bv_{t+1} = \bwt - \et\gt$ and $\E\gt = \bgt$. Notably, for any $t\geq 0$, there exists a $t_0 \leq t$, such that $t-t_0 \leq \tau-1$ and $\w_{t_0}^k = \bw_{t_0}$ for all $k = 1, 2, ..., N$. In this case, $\x\tk = \mathcal{S}_m(\w\tk)=\mathcal{S}(\w\tk-\bw_{t_0})+\bw_{t_0}$. Therefore, we have $\E \x\tk = \w\tk$ and $\E\Vert \x\tk - \w\tk \Vert^2 \leq q^2 \Vert\w\tk-\bw_{t_0} \Vert^2$.

\subsection{Key Lemmas}
To convey our proof clearly, it would be necessary to prove certain useful lemmas. We defer the proof of these lemmas to latter section and focus on proving the main theorem. 
\begin{lem}\label{lem:1}
(Results of one step SGD). Assume Assumption \ref{asmp:lsmooth} and \ref{asmp:ustrong}. If $\et \leq \frac{1}{4L}$, we have
\begin{equation}\label{eq:lem1}
\begin{aligned}
    \E\Vert \bv_{t+1}-\w^*\Vert ^2 
    &\leq (1-\et\mu) \E\Vert \bwt-\w^*\Vert ^2 + \et^2 \E\Vert \gt-\bgt\Vert ^2 +6L\et^2\Gamma + 2\sumkp \E\Vert \bwt - \x\tk \Vert ^2.
\end{aligned}
\end{equation}
\end{lem} 

\begin{lem}\label{lem:2}
(Bounding the variance). Assume Assumption \ref{asmp:gvaricane} holds. It follows that
\begin{equation}\label{eq:lem2}
\begin{aligned}
    \E\Vert \gt-\bgt\Vert ^2 \leq \sumk p_k^2\sigma^2.
\end{aligned}
\end{equation}
\end{lem} 

\begin{lem}\label{lem:3}
(Bounding the divergence of {$\x\tk$}). Assume Assumption \ref{asmp:gbound}, that $\et$ is non-increasing and $\et \leq 2\eta_{t+\tau}$ for all $t \geq 0$. It follows that
\begin{equation}\label{eq:lem3}
\begin{aligned}
    \sumkp\E\Vert \bwt-\x\tk\Vert ^2 \leq 4 (1+q^2) \et^2 (\tau-1)^2 G^2.
\end{aligned}
\end{equation}
\end{lem} 

\begin{lem}\label{lem:4}
(Bounding the divergence of {$\bwt$}). Assume Assumption \ref{asmp:gbound} and \ref{asmp:svariance}, that $\et$ is non-increasing and $\et \leq 2\eta_{t+\tau}$ for all $t \geq 0$. It follows that
\begin{equation}\label{eq:lem4}
\begin{aligned}
    \E\Vert \bw_{t+1} - \bv_{t+1} \Vert ^2 \leq 4 \sumk p_k^2 q^2 \et^2 \tau^2 G^2.
\end{aligned}
\end{equation}
\end{lem}

\subsection{Completing the Proof of Theorem \ref{thm:full}}
Note that $\E\bwt = \E\bvt$ when we take expectation to erase the randomness of stochastic binarization, therefore
\begin{equation}
\begin{aligned}\label{eq:thm_full_1}
    \E\Vert \bw_{t+1}-\w^*\Vert ^2 
    &= \E\Vert \bw_{t+1} - \bv_{t+1} + \bv_{t+1} -\w^* \Vert ^2 \\
    &= \E\Vert \bw_{t+1} - \bv_{t+1}\Vert ^2 + \E\Vert \bv_{t+1} -\w^* \Vert ^2.
\end{aligned}
\end{equation}
Let $\Delta_t = \E\Vert \bwt - \w^*\Vert^2$. From Lemma~\ref{lem:1}-\ref{lem:4}, it follows that
\begin{equation}
\begin{aligned}\label{eq:thm_full_2}
    \Delta_{t+1} \leq (1-\et\mu) \Delta_{t} + \et^2 B,
\end{aligned}
\end{equation}
where
\begin{equation}
\begin{aligned}
    B = \sumk p_k^2\sigma^2 + 6L\Gamma + 8(1+q^2)(\tau-1)^2G^2 + 4\sumk p_k^2q^2\tau^2G^2. 
\end{aligned}
\end{equation}
For a diminishing stepsize, $\et = \frac{\beta}{t+\gamma}$ for some $\beta > \frac{1}{\mu}$ and $\gamma > 0$ such that $\eta_1 \leq \min\{\frac{1}{\mu}, \frac{1}{4L}\}=\frac{1}{4L}$ and $\et\leq2\eta_{t+\tau}$. We will prove $\Delta_t\leq\frac{v}{t+\gamma}$ where $v = \max\{\frac{\beta^2B}{\beta\mu-1}, (\gamma+1)\Delta_1\}$. We prove it by induction. Firstly, the definition of $v$ ensures that it holds for $t = 1$. Assume the conclusion holds for some $t$, it follows that
\begin{equation}
\begin{aligned}
    \Delta_{t+1} 
    &\leq (1-\et\mu) \Delta_{t} + \et^2 B \\
    &\leq (1-\frac{\beta\mu}{t+\gamma}) \frac{v}{t+\gamma} + \frac{\beta^2B}{(t+\gamma)^2} \\
    &= \frac{t+\gamma-1}{(t+\gamma)^2}v + \frac{\beta^2B}{(t+\gamma)^2} - \frac{\beta\mu-1}{(t+\gamma)^2}v \\
    &\leq \frac{v}{t+\gamma+1}.
\end{aligned}
\end{equation}
Then by the $L$-smoothness of $F$,
\begin{equation}
\begin{aligned}
    \E[F(\bwt)]- F^* \leq \frac{L}{2} \Delta_t \leq  \frac{L}{2} \frac{v}{\gamma+t}.
\end{aligned}
\end{equation}
Specifically, if we choose $\beta = \frac{2}{\mu}, \gamma = \max\{8\frac{L}{\mu}, \tau\}-1$ and denote $\kappa = \frac{L}{\mu}$ , then $\et = \frac{2}{\mu}\frac{1}{\gamma+t}$. One can verify that the choice of $\et$ satisfies $\et\leq2\eta_{t+\tau}$ for $t \geq 1$. Then, we have
\begin{equation}
\begin{aligned}
    v=\max\{\frac{\beta^2B}{\beta\mu-1}, (\gamma+1)\Delta_1\}\leq\frac{\beta^2B}{\beta\mu-1} + (\gamma+1)\Delta_1 \leq\frac{4B}{\mu^2}+(\gamma+1)\Delta_1,
\end{aligned}
\end{equation}
and 
\begin{equation}
\begin{aligned}
    \E[F(\bwt)]- F^* \leq \frac{L}{2} \frac{v}{\gamma+t} \leq \frac{\kappa}{\gamma + t}(\frac{2B}{\mu}+\frac{\mu(\gamma+1)}{2}\Delta_1).
\end{aligned}
\end{equation}

\subsection{Deferred Proofs of Key Lemmas}
\textit{\textbf{Proof of Lemma~\ref{lem:1}.}} 
Notice that $\bv_{t+1} = \bwt - \et\gt$ and $\E\gt = \bgt$, then
\begin{equation}
\begin{aligned}\label{eq:lem1_1}
    \E\Vert \bv_{t+1}-\w^*\Vert ^2 
    &= \E\Vert \bwt - \et\gt - \w^* -\et\bgt + \et\bgt\Vert ^2 \\
    &= \underbrace{\E\Vert \bwt - \w^* - \et\bgt \Vert ^2}_{A_1} + \et^2 \E\Vert \gt - \bgt \Vert ^2.
\end{aligned}
\end{equation}
We next focus on bounding $A_1$. Again we split $A_1$ into three terms:
\begin{equation}
\begin{aligned}
    \Vert \bwt - \w^* - \et\bgt \Vert ^2 &= \Vert\bwt-\w^*\Vert ^2 \underbrace{-2\et \left< \bwt - \w^*, \bgt \right>}_{B_1} +\underbrace{\et^2\Vert \bgt \Vert ^2}_{B_2}.
\end{aligned}
\end{equation}
From the the L-smoothness of $F_k$, it follows that
\begin{equation}
\begin{aligned}\label{eq:l-smooth}
    \Vert \nabla F_k(\x\tk) \Vert ^2 \leq 2L(F_k(\x\tk)-F_k^*).
\end{aligned}
\end{equation}
By the convexity of $\Vert \cdot \Vert^2$ and Eq.~\ref{eq:l-smooth}, we have
\begin{equation}
\begin{aligned}
    B_2 = \et^2\Vert \bgt \Vert ^2\leq\et^2\sumkp \Vert \nabla F_k(\x\tk) \Vert^2 \leq 2L\et^2\sumkp(F_k(\x\tk)-F_k^*).
\end{aligned}
\end{equation}
Note that 
\begin{equation}
\begin{aligned}
    B_1 
    &= -2\et \left< \bwt - \w^*, \bgt \right> 
    = -2\et \sumkp \left< \bwt - \w^*, \nabla F_k(\x\tk) \right> \\
    &= -2\et \sumkp \left< \bwt - \x\tk, \nabla F_k(\x\tk) \right> -2\et \sumkp \left< \x\tk - \w^*, \nabla F_k(\x\tk) \right>.
\end{aligned}
\end{equation}
By Cauchy-Schwarz inequality and AM-GM inequality, we have
\begin{equation}
\begin{aligned}
    -2 \left< \bwt - \x\tk, \nabla F_k(\x\tk) \right> \leq \frac{1}{\et} \Vert \bwt - \x\tk \Vert^2 + \et \Vert \nabla F_k(\x\tk) \Vert^2.
\end{aligned}
\end{equation}
By the $\mu$-strong convexity of $F_k$, we have
\begin{equation}
\begin{aligned}
    -\left< \x\tk - \w^*, \nabla F_k(\x\tk) \right> \leq -(F_k(\x\tk) - F_k(\w^*)) -\frac{\mu}{2} \Vert \x\tk - \w^* \Vert^2.
\end{aligned}
\end{equation}
Therefore, we have
\begin{equation}
\begin{aligned}
    A_1 =\E\Vert \bwt - \w^* - \et\bgt \Vert ^2 
        &\leq \E\Vert\bwt-\w^*\Vert ^2 + 2L\et^2 \E\sumkp (F_k(\x\tk)-F_k^*) \\
        & + \et\E\sumkp (\frac{1}{\et} \Vert\bwt-\x\tk\Vert ^2 + \et\Vert \nabla F_k(\x\tk) \Vert ^2) \\
        & - 2\et\E\sumkp(F_k(\x\tk) - F_k(\w^*) + \frac{\mu}{2} \Vert \x\tk - \w^* \Vert^2) \\ 
        &\leq (1-\mu\et)\E\Vert\bwt-\w^*\Vert ^2 + \sumkp \E\Vert\bwt-\x\tk\Vert ^2 \\
        & + \tau[\underbrace{4L\et^2\sumkp (F_k(\x\tk )-F_k^*) - 2\et\sumkp(F_k(\x\tk) - F_k(\w^*))}_{C}]
\end{aligned}
\end{equation}
where we use Eq.\ref{eq:l-smooth} again and the inequality $-\E\Vert \x\tk - \w^* \Vert ^2 = -\E\Vert\bwt-\x\tk\Vert ^2 - \E\Vert\bwt-\w^*\Vert ^2 \leq -\E\Vert\bwt-\w^*\Vert ^2$.

We next aim to bound $C$. We define $\gmt = 2\et(1-2L\et)$. Since $\et\leq\frac{1}{4L} , \et \leq \gmt \leq 2\et$. Then we split $C$ into two terms:
\begin{equation}
\begin{aligned}
    C &= -2\et(1-2L\et) \sumkp (F_k(\x\tk )-F_k^*) + 2\et \sumkp (F_k(\w^*) - F_k^*) \\
    &= -\gmt \sumkp (F_k(\x\tk )-F^*) + (2\et-\gmt) \sumkp (F^* - F_k^*) \\
    &= \underbrace{-\gmt \sumkp (F_k(\x\tk )-F^*)}_{D} + 4L\et^2\Gamma \\
\end{aligned}
\end{equation}
where in the last equation, we use the notation $\Gamma = \sumkp (F^*-F_k^*) = F^*-\sumkp F_k^*$. 

To bound $D$, we have
\begin{equation}
\begin{aligned}
    \sumkp (F_k(\x\tk )-F^*) 
    &= \sumkp (F_k(\x\tk )-F_k(\bwt)) + \sumkp (F_k(\bwt )-F^*) \\
    &\geq \sumkp \left< \nabla F_k(\bwt), \x\tk-\bwt\right> + F(\bwt) - F^* \\
    &\geq -\frac{1}{2}\sumkp [\et \Vert \nabla F_k(\bwt)\Vert^2 + \frac{1}{\et}\Vert \x\tk-\bwt \Vert^2 ] + F(\bwt) - F^* \\
    &\geq -\sumkp [\et L(F_k(\bwt)-F_k^*) + \frac{1}{2\et}\Vert \x\tk-\bwt \Vert^2 ] + F(\bwt) - F^* \\
\end{aligned}
\end{equation}
where the first inequality results from the convexity of $F_k$, the second inequality from AM-GM inequality and the third inequality from Eq.~\ref{eq:l-smooth}. Therefore
\begin{equation}
\begin{aligned}
    C
    & = \gmt\sumkp [\et L(F_k(\bwt)-F_k^*) + \frac{1}{2\et}\Vert \x\tk-\bwt \Vert^2] - \gmt(F(\bwt) - F^*) +4L\et^2\Gamma  \\
    & = \gmt(\et L-1)\sumkp (F_k(\bwt)-F_k^*)  +(4L\et^2+\gmt\et L)\Gamma + \frac{\gmt}{2\et}\sumkp \Vert \x\tk-\bwt \Vert^2 \\
    & \leq 6L\et^2 \Gamma + \sumkp \Vert \x\tk-\bwt \Vert^2 \\
\end{aligned}
\end{equation}
where in the last inequality, we use the following facts: (1)$\et L-1\leq-\frac{3}{4}\leq 0$ and $\sumkp (F_k(\bwt)-F^*) = F(\bwt)-F^* \geq 0$ (2) $\Gamma \geq 0$ and $4L\et^2  + \gmt\et L \leq 6\et^2 L$ and (3) $\frac{\gmt}{2\et} \leq 1$.

Recalling the expression of $A_1$ and plugging $C$ into it, we have
\begin{equation}
\begin{aligned}
    A_1
    & = \E\Vert \bwt-\w^*-\et\bgt \Vert^2 
     \leq (1-\mu\et)\E\Vert\bwt-\w^*\Vert ^2 + 2\sumkp \E\Vert\bwt-\x\tk\Vert ^2 + 6L\et^2 \Gamma.
\end{aligned}
\end{equation}
Plugging $A_1$ into Eq.~\ref{eq:lem1_1}, we have the result in Lemma~\ref{lem:1}
\begin{equation}
\begin{aligned}
    \E\Vert \bv_{t+1}-\w^*\Vert ^2 
    &\leq (1-\et\mu) \E\Vert \bwt-\w^*\Vert ^2 + \et^2 \E\Vert \gt-\bgt\Vert ^2 +6L\et^2\Gamma + 2\sumkp \E\Vert \bwt - \x\tk \Vert ^2.
\end{aligned}
\end{equation}

\textit{\textbf{Proof of Lemma~\ref{lem:2}.}}
From Assumption \ref{asmp:gvaricane}, the variance of the stochastic gradients in device $k$ is bounded by $\sigma^2$, then
\begin{equation}
\begin{aligned}
    \E\Vert \gt-\bgt \Vert ^2 
    & = \E \Vert \sumkp (\nabla F_k(\w\tk, \xi\tk) - \nabla F_k(\w\tk)) \Vert ^2 \\
    & = \sumk p_k^2\E \Vert \nabla F_k(\w\tk, \xi\tk) - \nabla F_k(\w\tk) \Vert ^2 \\
    & \leq \sumk p_k^2 \sigma^2 \\
\end{aligned}
\end{equation}

\textit{\textbf{Proof of Lemma~\ref{lem:3}.}}
Considering that $\E \x\tk = \w\tk$, we have
\begin{equation}
\begin{aligned}\label{eq:lem3_1}
    \sumkp\E\Vert \bwt-\x\tk\Vert ^2 
    &= \sumkp\E\Vert (\bwt-\w\tk) - (\w\tk-\x\tk)\Vert ^2 \\
    &= \sumkp\E\Vert \bwt-\w\tk \Vert ^2  + \sumkp\E\Vert \w\tk-\x\tk \Vert ^2 
\end{aligned}
\end{equation}
Since FedBAT requires a communication each $\tau$ steps. Therefore, for any $t\geq 0$, there exists a $t_0 \leq t$, such that $t-t_0 \leq \tau-1$ and $\w_{t_0}^k = \bw_{t_0}$ for all $k = 1, 2, ..., N$. Also, we use the fact that $\et$ is non-increasing and $\eta_{t_0} \leq 2\et$ for all $t-t_0 \leq \tau-1$, then
\begin{equation}
\begin{aligned}\label{eq:lem3_2}
    \sumkp\E\Vert \bwt-\w\tk \Vert ^2
    &= \sumkp\E\Vert (\w\tk-\bw_{t_0}) - (\bwt-\bw_{t_0}) \Vert ^2 \\
    &\leq \sumkp\E\Vert \w\tk-\bw_{t_0}\Vert ^2 \\
    &\leq \sumkp \E\sum_{t=t_0}^{t-1}(\tau-1)\et^2\Vert \nabla F_k(\x\tk, \xi\tk)\Vert ^2 \\
    &\leq \sumkp \sum_{t=t_0}^{t-1}(\tau-1)\et^2 G^2 \\
    &\leq \sumkp \et^2(\tau-1)^2 G^2 \\
    &\leq 4 \et^2(\tau-1)^2 G^2 \\
\end{aligned}
\end{equation}
Here in the first inequality, we use $\E\Vert X-\E X \Vert^2 \leq \E\Vert X \Vert^2$ where $X = \w\tk - \bw_{t_0}$ with probability $p_k$. In the second inequality, we use Jensen inequality:
\begin{equation}
\begin{aligned}\label{eq:lem3_3}
    \Vert \w\tk-\bw_{t_0}\Vert ^2 = \Vert \sum_{t=t_0}^{t-1} \et \nabla F_k(\w\tk, \xi\tk) \Vert ^2 
    \leq (t-t_0)\sum_{t=t_0}^{t-1} \et^2 \Vert \nabla F_k(\w\tk, \xi\tk) \Vert ^2.
\end{aligned}
\end{equation}
In the third inequality, we use $\et \leq \eta_{t_0}$ for $t \geq t_0$ and $\E \Vert \nabla F_k(\w\tk, \xi\tk) \Vert ^2 \leq G^2$ for $k = 1, 2, ..., N$ and $t\geq 1$. In the last inequality, we use $\eta_{t_0}\leq 2 \eta_{t_0+\tau} \leq 2\et$ for $t_0 \leq t \leq t_0 + \tau$.

According to Assumption \ref{asmp:svariance}, we have $\E\Vert \w\tk-\x\tk \Vert ^2 \leq q^2 \E\Vert \w\tk-\bw_{t_0} \Vert ^2$ as discussed in Section~\ref{sec:c1}. Then the second term in Eq.~\ref{eq:lem3_1} can be bounded by reusing the result in Eq.~\ref{eq:lem3_2} as
\begin{equation}
\begin{aligned}\label{eq:lem3_4}
    \sumkp\E\Vert \w\tk-\x\tk \Vert ^2 \leq  q^2 \sumkp \E\Vert \w\tk-\bw_{t_0} \Vert ^2 \leq 4q^2 \et^2(\tau-1)^2 G^2. 
\end{aligned}
\end{equation}
Plugging Eq.~\ref{eq:lem3_2} and Eq.~\ref{eq:lem3_4} into Eq.~\ref{eq:lem3_1}, we have the result in Lemma~\ref{lem:3}
\begin{equation}
\begin{aligned}
    \sumkp\E\Vert \bwt-\x\tk\Vert ^2 \leq 4 (1+q^2) \et^2 (\tau-1)^2 G^2.
\end{aligned}
\end{equation}

\textit{\textbf{Proof of Lemma~\ref{lem:4}.}}
Notice that $\bw_{t+1}=\bv_{t+1}$ when $t+1 \notin \ie$ and $\bw_{t+1}=\sumkp \mathcal{S}_m(\v_{t+1}), \bv_{t+1}=\sumkp \v_{t+1}$ when $t+1 \in \ie$. Hence, if $t+1 \in \ie$, we have
\begin{equation}
\begin{aligned}\label{eq:lem4_1}
    \E\Vert \bw_{t+1} - \bv_{t+1} \Vert ^2 
    &= \E\Vert \sumkp (\mathcal{S}_m(\v\ttk) - \v\ttk)\Vert ^2 \\
    &= \sumk p_k^2 \E\Vert (\mathcal{S}_m(\v\ttk) - \v\ttk)\Vert ^2 \\
    &\leq \sumk p_k^2 q^2 \E\Vert \v\ttk - \bw_{t_0}\Vert ^2 \\
    &\leq 4 \sumk p_k^2  q^2 \et^2 \tau^2 G^2, 
\end{aligned}
\end{equation}
where the last inequality follows the result in Eq.\ref{eq:lem3_3}.

\section{Proof of Theorem \ref{thm:part}}\label{sec:proof_part}
In this section, we analyze FedBAT in the setting of partial device participation. 

\subsection{Additional Notation}\label{sec:d1}
Recall that $\w\tk$ is the model parameter maintained in the $k$-th device at the $t$-th step. $\ie = \{n\tau | n = 1, 2, ..., N\}$ is the set of global synchronization steps. Again, $\bgt = \sumkp \nabla F_k(\x\tk)$ and $\gt = \sumkp \nabla F_k(\x\tk, \xi\tk)$. Therefore, $\bv_{t+1} = \bwt - \et\gt$ and $\E\gt = \bgt$.

Now we consider the case where FedBAT samples a random set $\S_t$ of devices to participate in each round of training. This make the analysis a little bit intricate, since $\S_t$ varies each $\tau$ steps. Following~\cite{fedavg_convergence}, we assume that FedBAT always activates all devices at the beginning of each round and then uses the parameters maintained in only a few sampled devices to produce the next-round parameter. It is clear that this updating scheme is equivalent to the original. As assumed in Theorem~\ref{thm:part} that $p_1=...=p_N=\frac{1}{N}$, the update of FedBAT with partial devices active can be described as: for all $k\in [N]$,
\begin{equation}
\begin{aligned}
    \v\ttk = \w\tk - \et \nabla F_k(\x\tk, \xi\tk)
\end{aligned}
\end{equation}
\begin{equation}
\begin{aligned}
    \x_{t}^{k} = \mathcal{S}_m(\w\tk)
\end{aligned}
\end{equation}
\begin{equation}
\begin{aligned}
\w\ttk = \left\{
\begin{array}{ll}
    \v\ttk          & \text{if} \ t+1 \notin \ie,\\ 
    \frac{\sum_{k\in \S_{t+1}} p_k \mathcal{S}_m(\v\tk)}{\sum_{k\in \S_{t+1}} p_k} = \frac{1}{K} \sum_{k\in \S_{t+1}} \mathcal{S}_m(\v\ttk)  & \text{if} \ t+1 \in    \ie. \\
\end{array}
\right. \\
\end{aligned}
\end{equation}

\subsection{Key Lemmas}
\begin{lem}\label{lem:5}
(Unbiased sampling scheme). In the case of partial device participation in Theorem~\ref{thm:part}, we have
\begin{equation}\label{eq:lem5}
\begin{aligned}
    \E \bw_{t+1} = \E \bv_{t+1}. 
\end{aligned}
\end{equation}
\end{lem} 

\begin{lem}\label{lem:6}
(Bounding the variance of $\bwt$). In the case of partial device participation in Theorem~\ref{thm:part}, with Assumption \ref{asmp:gbound} and \ref{asmp:svariance}, assume that $\et$ is non-increasing and $\et \leq \eta_{t+\tau}$ for all $t \geq 0$. It follows that
\begin{equation}\label{eq:lem6}
\begin{aligned}
    \E\Vert \bw_{t+1} - \bv_{t+1} \Vert ^2 \leq 4\frac{q^2(N-1)+N-K}{K(N-1)}\et^2\tau^2G^2.
\end{aligned}
\end{equation}
\end{lem}

\subsection{Completing the Proof of Theorem \ref{thm:part}}
Using Lemma~\ref{lem:5}, Eq.\ref{eq:thm_full_1} still holds, that is
\begin{equation}
\begin{aligned}
    \E\Vert \bw_{t+1}-\w^*\Vert ^2 
    &= \E\Vert \bw_{t+1} - \bv_{t+1} + \bv_{t+1} -\w^* \Vert ^2 \\
    &= \E\Vert \bw_{t+1} - \bv_{t+1}\Vert ^2 + \E\Vert \bv_{t+1} -\w^* \Vert ^2 \\
\end{aligned}
\end{equation}
Let $\Delta_t = \E\Vert \bwt - \w^*\Vert^2$. From Lemma~\ref{lem:1}, \ref{lem:2}, \ref{lem:3} and \ref{lem:6}, it follows that
\begin{equation}
\begin{aligned}\label{eq:thm_part_1}
    \Delta_{t+1} \leq (1-\et\mu) \Delta_{t} + \et^2 B
\end{aligned}
\end{equation}
where
\begin{equation}
\begin{aligned}
    B=\frac{\sigma^2}{N} + 6L\Gamma + 8(1+q^2)(\tau-1)^2G^2 + 4\frac{q^2(N-1)+N-K}{K(N-1)}\tau^2G^2
\end{aligned}
\end{equation}
The only difference between Eq.\ref{eq:thm_part_1} and Eq.\ref{eq:thm_full_2} is the value of constant $B$. Following the same process, we can get the result of Theorem~\ref{thm:part}
\begin{equation}
\begin{aligned}
    \E[F(\bwt)]- F^* \leq \frac{L}{2} \frac{v}{\gamma+t} \leq \frac{\kappa}{\gamma + t}(\frac{2B}{\mu}+\frac{\mu(\gamma+1)}{2}\Vert\w_1-\w^*\Vert^2),
\end{aligned}
\end{equation}
where $\gamma = \max\{8\frac{L}{\mu}, \tau\}-1$ and $\kappa = \frac{L}{\mu}$.

\subsection{Deferred Proofs of Key Lemmas}
\textit{\textbf{Proof of Lemma~\ref{lem:5}.}}
Considering the partial device participation in Theorem~\ref{thm:part}, $\bw_{t+1}=\bv_{t+1}$ when $t+1 \notin \ie$ and $\bw_{t+1}=\frac{1}{K} \sum_{k\in \S_{t+1}} \mathcal{S}_m(\v\ttk), \bv_{t+1}=\frac{1}{N} \sumk \v\ttk$ when $t+1 \in \ie$. In the latter case, there are two kinds of randomness between $\bw_{t+1}$ and $\bv_{t+1}$, respectively from the client's random selection and stochastic binarization. To distinguish them, we use the notation $\E_{\S_t}$ when we take expectation to erase the randomness of device selection, and use the notation $\E_{\mathcal{S}}$ when we take expectation to erase the randomness of binarization. Therefore, when $t+1 \in \ie$, we have
\begin{equation}
\begin{aligned}
    \E\bw_{t+1} = \E_{\S_t}[\E_{\mathcal{S}}\bw_{t+1}] = \E_{\S_t}[\E_{\mathcal{S}}\frac{1}{K} \sum_{k\in \S_{t+1}} \mathcal{S}_m(\v\ttk)] = \E_{\S_t}[ \frac{1}{K} \sum_{k\in \S_{t+1}} \v\ttk] = \bv_{t+1}
\end{aligned}
\end{equation}

\textit{\textbf{Proof of Lemma~\ref{lem:6}.}} Notice that $\bw_{t+1}=\bv_{t+1}$ when $t+1 \notin \ie$ and $\bw_{t+1}=\frac{1}{K} \sum_{k\in \S_{t+1}} \mathcal{S}_m(\v\ttk), \bv_{t+1}=\sumkp \v_{t+1}$ when $t+1 \in \ie$. Hence, we have
\begin{equation}
\begin{aligned}
    \E\Vert \bw_{t+1} - \bv_{t+1} \Vert ^2 
    &= \E\Vert \frac{1}{K} \sum_{k\in \S_{t+1}} \mathcal{S}_m(\v\ttk) - \bv_{t+1}\Vert ^2 \\
    &= \E\Vert \frac{1}{K} \sum_{k\in \S_{t+1}} (\mathcal{S}_m(\v\ttk) - \v\ttk) + \frac{1}{K} \sum_{k\in \S_{t+1}} \v\ttk - \bv_{t+1}\Vert ^2 \\
    &= \underbrace{\E\Vert \frac{1}{K} \sum_{k\in \S_{t+1}} (\mathcal{S}_m(\v\ttk) - \v\ttk) \Vert ^2}_{A_1} + \underbrace{\E\Vert \frac{1}{K} \sum_{k\in \S_{t+1}} \v\ttk - \bv_{t+1}\Vert ^2}_{A_2} \\
\end{aligned}
\end{equation}
To bound $A_1$, we have
\begin{equation}
\begin{aligned}
    A_1
    &= \E\Vert \frac{1}{K} \sum_{k\in \S_{t+1}} (\mathcal{S}_m(\v\ttk) - \v\ttk) \Vert ^2 \\
    &= \frac{1}{K^2} \sum_{k\in \S_{t+1}} \E\Vert  (\mathcal{S}_m(\v\ttk) - \v\ttk) \Vert ^2 \\
    &\leq \frac{1}{K^2} \sum_{k\in \S_{t+1}} 4q^2\et^2\tau^2G^2 \\
    &= \frac{4}{K} q^2\et^2\tau^2G^2,
\end{aligned}
\end{equation}
where in the inequality we use the result of Eq.\ref{eq:lem4_1}. Then, to bound $A_2$, we have
\begin{equation}
\begin{aligned}
    A_2 
    &= \E\Vert \frac{1}{K} \sum_{k\in \S_{t+1}} \v\ttk - \bv_{t+1}\Vert ^2 \\
    &=  \frac{1}{K^2} \E\Vert \sum_{i=1}^N \mathbb{I}\{i\in\S_{t+1}\}(\v_{t+1}^i-\bv_{t+1})\Vert ^2 \\
    &=  \frac{1}{K^2} \E[\sum_{i=1}^N \mathbb{P}(i\in\S_{t+1}) \Vert \v_{t+1}^i-\bv_{t+1}\Vert ^2 + \sum_{i \neq j} \mathbb{P}(i, j\in\S_{t+1}) \left< \v_{t+1}^i-\bv_{t+1}, \v_{t+1}^j-\bv_{t+1} \right>]\\
    &= \frac{1}{KN} \E\sum_{i=1}^N \Vert \v_{t+1}^i-\bv_{t+1}\Vert ^2 + \frac{K-1}{KN(N-1)}\E\sum_{i \neq j} \left< \v_{t+1}^i-\bv_{t+1}, \v_{t+1}^j-\bv_{t+1} \right> \\ 
    &= \frac{N-K}{KN(N-1)} \sum_{i=1}^N \E\Vert \v_{t+1}^i-\bv_{t+1}\Vert ^2   \\ 
\end{aligned}
\end{equation}
where we use the following equalities: (1) $\mathbb{P}(i\in\S_{t+1}) = \frac{K}{N}$ and $\mathbb{P}(i, j\in\S_{t+1}) = \frac{K(K-1)}{N(N-1)}$ for all $i\neq j$ and (2) $\sum_{i=1}^N \Vert \v_{t+1}^i-\bv_{t+1}\Vert ^2 + \sum_{i \neq j} \left< \v_{t+1}^i-\bv_{t+1}, \v_{t+1}^j-\bv_{t+1} \right> = 0$. Also using the result of Eq.\ref{eq:lem4_1}, we have
\begin{equation}
\begin{aligned}
    A_2 
    &= \frac{N-K}{KN(N-1)} \sum_{i=1}^N \E\Vert \v_{t+1}^i-\bv_{t+1}\Vert ^2   \\ 
    &\leq \frac{N-K}{K(N-1)} 4\et^2\tau^2G^2   \\ 
\end{aligned}
\end{equation}
Plugging $A_1$ and $A_2$, we have the result in Lemma~\ref{lem:6}
\begin{equation}
\begin{aligned}
    \E\Vert \bw_{t+1} - \bv_{t+1} \Vert ^2  \leq 4\frac{q^2(N-1)+N-K}{K(N-1)}\et^2\tau^2G^2.
\end{aligned}
\end{equation}

\section{Proof of Theorem \ref{thm:non-convex}}
In this section, we proof the convergence of FedBAT under non-convex settings. We only use 
Assumptions \ref{asmp:lsmooth}, \ref{asmp:gvaricane}, \ref{asmp:gbound}, \ref{asmp:svariance}. Note that Assumptions \ref{asmp:ustrong} is about the strongly convex, which we do not use in this case. 

\subsection{Additional Notation}
Let us review the notations in Section \ref{sec:d1}. $\w_t^k$ is the model parameters maintained in the $k$-th device at the $t$-th step. Let $\ie$ be the set of global synchronization steps, i.e., $\ie = \{n\tau | n = 1, 2, ...\}$. If $t+1\in\ie$, i.e., the time step to communication, FedBAT activates all devices. Then the update of FedBAT can be described as

\begin{equation}
\begin{aligned}
    \v\ttk = \w\tk - \et \nabla F_k(\x\tk, \xi\tk)
\end{aligned}
\end{equation}
\begin{equation}
\begin{aligned}
    \x_{t}^{k} = \mathcal{S}_m(\w\tk)
\end{aligned}
\end{equation}
\begin{equation}
\begin{aligned}
\w\ttk = \left\{
\begin{array}{ll}
    \v\ttk          & \text{if} \ t+1 \notin \ie,\\ 
    \sumkp \mathcal{S}_m(\v\ttk)   & \text{if} \ t+1 \in    \ie. \\
\end{array}
\right. \\
\end{aligned}
\end{equation}

Here, the variable $\v\ttk$ is introduced to represent the immediate result of one step SGD update from $\w\tk$. We interpret $\w\ttk$ as the parameter obtained after communication steps (if possible). 
Also, an additional variable $\x\tk$ is introduced to represent the result of binarization on model update. 

In our analysis, we define two virtual sequences $\bvt=\sumkp \v\tk$ and $\bwt=\sumkp \w\tk$. It is obviously that $\E\bvt = \E\bwt$. $\bv_{t+1}$ results from an single step of SGD from $\bwt$. When $t+1 \notin \ie$, both are inaccessible. When $t+1 \in \ie$, we can only fetch $\bw_{t+1}$. For convenience, we define $\bgt = \nabla F(\x\tk) =\sumkp \nabla F_k(\x\tk)$ and $\gt = \sumkp \nabla F_k(\x\tk, \xi\tk)$. Therefore, $\bv_{t+1} = \bwt - \et\gt$ and $\E\gt = \bgt$. Notably, for any $t\geq 0$, there exists a $t_0 \leq t$, such that $t-t_0 \leq \tau-1$ and $\w_{t_0}^k = \bw_{t_0}$ for all $k = 1, 2, ..., N$. In this case, $\x\tk = \mathcal{S}_m(\w\tk)=\mathcal{S}(\w\tk-\bw_{t_0})+\bw_{t_0}$. Therefore, we have $\E \x\tk = \w\tk$ and $\E\Vert \x\tk - \w\tk \Vert^2 \leq q^2 \Vert\w\tk-\bw_{t_0} \Vert^2$.

\subsection{Key Lemmas}

\begin{lem}\label{lem:7}
Assume Assumption \ref{asmp:lsmooth} and \ref{asmp:gvaricane}, we have
\begin{equation}
\begin{aligned}\label{eq:lem_8}
    \E F(\bw_{t+1})
    &\leq \E F(\bwt) -\frac{\eta}{2} \E \Vert \nabla F(\bwt)\Vert ^2 
    +\frac{L\eta^2-\eta}{2}\E\Vert \bgt\Vert^2 \\
    & \quad 
    +\frac{\eta}{2}L^2\frac{1}{N}\sumk\E\Vert \bwt - \x\tk \Vert^2 
    + \frac{L\eta^2}{2}\E\Vert \bgt -\gt \Vert^2  + \frac{L}{2} \E \Vert \bw_{t+1}- \bv_{t+1} \Vert^2\\
\end{aligned}
\end{equation}
\end{lem} 

\subsection{Completing the Proof of Theorem \ref{thm:non-convex}}
Since $\eta = \frac{1}{L\sqrt{T}}$, we have $\frac{L\eta^2-\eta}{2} \leq 0$, then Eq.(\ref{eq:lem_8}) can be rewritten as
\begin{equation}
\begin{aligned}
    \E F(\bw_{t+1})
    &\leq \E F(\bwt) -\frac{\eta}{2} \E \Vert \nabla F(\bwt)\Vert ^2 
    +\frac{\eta}{2}L^2\frac{1}{N}\sumk\E\Vert \bwt - \x\tk \Vert^2 
    + \frac{L\eta^2}{2}\E\Vert \bgt -\gt \Vert^2 + \frac{L}{2} \E \Vert \bw_{t+1}- \bv_{t+1} \Vert^2\\
\end{aligned}
\end{equation}
The last three terms can be bounded by Lemmas \ref{lem:2}, \ref{lem:3} and  \ref{lem:6}. Note that these lemmas require no convex assumption. Therefore,
\begin{equation}
\begin{aligned}
    \E F(\bw_{t+1})
    &\leq \E F(\bwt) -\frac{\eta}{2} \E \Vert \nabla F(\bwt)\Vert ^2 \\
    &\quad 
    +\frac{\eta}{2}L^2 4(1+q^2)\eta^2(\tau-1)^2G^2
    + \frac{L\eta^2}{2} \frac{\sigma^2}{N}
+ \frac{L}{2} 4\frac{q^2(N-1)+N-K}{K(N-1)}\et^2\tau^2G^2
\end{aligned}
\end{equation}

Now rearranging  the terms and summing over $t=0,...,T-1$ yield that
\begin{equation}
\begin{aligned}
    \frac{1}{2}\eta \frac{1}{T}\sum_{t=0}^{T-1}\E\Vert \nabla F(\bwt)\Vert^2 
    &\leq \frac{F(\w_0) - F(\w_T)}{T} \\
    &\quad 
    +\frac{\eta}{2}L^2 4(1+q^2)\eta^2(\tau-1)^2G^2
    + \frac{L\eta^2}{2} \frac{\sigma^2}{N}
+ \frac{L}{2} 4\frac{q^2(N-1)+N-K}{K(N-1)}\et^2\tau^2G^2
\end{aligned}
\end{equation}
Picking the learning rate $\eta=\frac{1}{L\sqrt{T}}$, and with $F(\w_T)\geq \frac{1}{N}\sum_{k=1}^N F_k^* = F^* - \Gamma$, we have
\begin{equation}
\begin{aligned}
    \frac{1}{T} \sum_{t=0}^{T-1}\E\Vert \nabla F(\bwt)\Vert^2 
    &\leq \frac{2L(F(\bw_0) - F^* + \Gamma)}{\sqrt{T}} + \frac{P}{\sqrt{T}} + \frac{Q}{T},
\end{aligned}
\end{equation}
where $P = \frac{\sigma^2}{N} + 4\frac{q^2(N-1)+N-K}{K(N-1)}\tau^2G^2$ and $Q=4(1+q^2)(\tau-1)^2G^2$.

\subsection{Deferred Proofs of Key Lemmas}
\textit{\textbf{Proof of Lemma~\ref{lem:7}.}}
For any $L$-smooth function $F$, we have 
\begin{equation}
    F(\bw_{t+1}) \leq F(\bv_{t+1}) + \left< \nabla F(\bv_{t+1}), \bw_{t+1} -\bv_{t+1}\right> +\frac{L}{2} \Vert \bw_{t+1} -\bv_{t+1} \Vert^2
\end{equation}
As $\E\bw_{t+1} =\E\bv_{t+1}$, taking expectations for the randomness of stochastic binarization and client selection yields that
\begin{equation}\label{eq:lem7_1}
    \E F(\bw_{t+1}) \leq \E F(\bv_{t+1}) + \frac{L}{2} \E \Vert \bw_{t+1} -\bv_{t+1} \Vert^2
\end{equation}
Since $\bv_{t+1}=\bwt - \eta \gt$, with $L$-smoothness, we have 
\begin{equation}
    F(\bv_{t+1}) \leq F(\bwt) -\eta \left< \nabla F(\bwt), \gt\right> +\frac{L\eta^2}{2} \Vert \gt \Vert^2
\end{equation}
The inner product term above can be written in expectation as follows:
\begin{equation}\label{eq:lem7_2}
    2\E \left< \nabla F(\bwt), \gt\right> = \E \Vert \nabla F(\bwt)\Vert ^2+\E\Vert \gt\Vert^2 -\E \Vert \nabla F(\bwt)-\gt \Vert^2
\end{equation}

Now, we consider the last term in Eq.\ref{eq:lem7_2} with the fact that $\E\gt= \E\bgt$
\begin{equation}
\begin{aligned}
    \E\Vert \nabla F(\bwt)-\gt \Vert^2 
    & = \E\Vert \nabla F(\bwt)-\bgt + \bgt -\gt \Vert^2 \\
    & = \E\Vert \nabla F(\bwt)-\bgt\Vert^2 + \E\Vert \bgt -\gt \Vert^2 \\
    &= \E\Vert \frac{1}{N}\sumk (\nabla F_k(\bwt) - \nabla F_k(\x\tk)) \Vert^2 + \E\Vert \bgt -\gt \Vert^2 \\
    &\leq L^2 \frac{1}{N}\sumk \E\Vert \bwt - \x\tk \Vert^2 +  \E\Vert \bgt -\gt \Vert^2 \\
\end{aligned}
\end{equation}
Further, we have
\begin{equation}\label{eq:lem7_3}
\begin{aligned}
    -\eta \E \left< \nabla F(\bwt), \gt\right>
    &\leq -\frac{\eta}{2} \E \Vert \nabla F(\bwt)\Vert ^2-\frac{\eta}{2}\E\Vert \gt\Vert^2 +\frac{\eta}{2}L^2\frac{1}{N}\sumk\E \Vert \bwt - \x\tk \Vert^2 + \frac{\eta}{2}\E \Vert \bgt -\gt \Vert^2 \\
\end{aligned}
\end{equation}
Summing Eq.\ref{eq:lem7_3} into Eq.\ref{eq:lem7_2}, we have
\begin{equation}
\begin{aligned}
    \E F(\bv_{t+1}) \leq \E F(\bwt) -\frac{\eta}{2} \E \Vert \nabla F(\bwt)\Vert ^2 +(\frac{L\eta^2}{2}-\frac{\eta}{2})\E\Vert \gt\Vert^2 +\frac{\eta}{2}L^2\frac{1}{N}\sumk\E \Vert \bwt - \x\tk \Vert^2 + \frac{\eta}{2}\E \Vert \bgt -\gt \Vert^2 \\
\end{aligned}
\end{equation}

$\E\Vert \gt\Vert^2$ can be expanded as follows:
\begin{equation}
    \E\Vert \gt\Vert^2 = \E\Vert \gt - \bgt + \bgt \Vert^2 = \E\Vert \bgt \Vert^2 +\E\Vert \gt - \bgt \Vert^2
\end{equation}
Therefore, we have
\begin{equation}
\begin{aligned}\label{eq:lem7_4}
    \E F(\bv_{t+1}) 
    &\leq \E F(\bwt) -\frac{\eta}{2} \E \Vert \nabla F(\bwt)\Vert ^2 
    +(\frac{L\eta^2}{2}-\frac{\eta}{2})\E\Vert \bgt\Vert^2 
    +\frac{\eta}{2}L^2\frac{1}{N}\sumk\E \Vert \bwt - \x\tk \Vert^2 
    + \frac{L\eta^2}{2}\E \Vert \bgt -\gt \Vert^2 \\
\end{aligned}
\end{equation}
Finally, summing Eq.\ref{eq:lem7_4} into Eq.\ref{eq:lem7_1} yields the result in Lemma~\ref{lem:7}.

\end{document}